\def\gmode{} %%use this to include figures

 \def\gdriver{dviout}%for dviout for preview

\newcommand{\dtitle}[1]{\title{ \if \gmode \else
\color{red} Demo mode!\\
comment out \textbackslash def \textbackslash gmode\{demo\} at the header to include figures \color{black}\\
\fi
#1 }}
\ifx\pdfoutput\undefined
\else
 \def\gdriver{}
\fi
 \documentclass[\gmode,\gdriver,runningheads]{llncs}

\usepackage{amsmath,amssymb} % define this before the line numbering.
\usepackage{color}

\usepackage{caption}
\usepackage{subcaption}
\usepackage{booktabs}
\usepackage{rotating, multirow}

\graphicspath{{./figs/}}

\newcommand{\fnote}[1]{}

\newcommand{\mnote}[1]{}
\newcommand{\kcut}[1]{}
\newcommand{\jptext}[1]{}
\ifx\pdfoutput\undefined
\else
 \usepackage{CJKutf8}
 \renewcommand{\fnote}[1]{}
 \renewcommand{\mnote}[1]{}
 \renewcommand{\jptext}[1]{}
\fi

\newcommand{\etal}{\textit{et al}.}
\newcommand{\ie}{\textit{i}.\textit{e}.}

\begin{document}

%macro for raising the point in decimal numbers; see example in the abstract
\newcommand{\point}{
    \raise0.7ex\hbox{.}
    }

%Do   -- NOT --    use any additional macros

\pagestyle{headings}

\mainmatter

%\def\ACCV16SubNumber{577}  % Insert your submission number here

%===========================================================
\dtitle{Simultaneous independent image display technique on multiple 3D objects}

\titlerunning{Simultaneous independent image display technique on multiple 3D objects}
\authorrunning{T. Hirukawa, M. Visentini-Scarzanella, H. Kawasaki, R. Furukawa, S. Hiura}

\author{Takuto Hirukawa\inst{1} \and Marco Visentini-Scarzanella\inst{1} \and Hiroshi Kawasaki\inst{1} \and \\Ryo Furukawa\inst{2} \and Shinsaku Hiura\inst{2}}
\institute{Computer Vision and Graphics Laboratory,\\
Kagoshima University, Japan\\
\and Graduate School of Information Sciences,\\
Hiroshima City University, Japan}

\maketitle

%===========================================================
\begin{abstract}
We propose a new system to visualize depth-dependent patterns and images on solid objects with complex geometry using multiple projectors. The system, despite consisting of conventional passive LCD projectors, is able to project different images and patterns depending on the spatial location of the object. The technique 
is based on the simple principle that multiple patterns projected from multiple 
projectors interfere constructively with each other when their patterns are projected on the same object.
Previous techniques based on the same principle can only achieve 1) low resolution 
    volume colorization or 2) high resolution images but only on a limited number 
    of flat planes. In this paper, we discretize a 3D object into a number of 3D points so that high resolution images can be projected onto the complex 
    shapes. We also propose a dynamic ranges expansion technique as well as 
    an efficient optimization procedure based on epipolar constraints. 
%limitations found in previous systems. 
Such technique can be used to the extend projection mapping to have spatial dependency, which is desirable for practical applications. We also demonstrate the system potential as a visual instructor for object placement and assembling. Experiments prove the effectiveness of 
    our method.
%\dots
\end{abstract}

%===========================================================
\section{Introduction}
\vspace{-0.1cm}
%\knote{[Kawasaki] Should rewrite. Our purpose is to project indendent images to multiple 3D 
%shapes, which is a simple extension of our previous PSIVT paper. One promising 
%application is for construction purpose; certainly Lena or Mandorill images are 
%not suitable for the purpose.}
%\knote{[Kawasaki] describe the difference from [5] more clearly.}

Recent improvements in projectors' resolution, brightness and cost 
effectiveness, extended their pervasiveness beyond flat screen projection, and to projection mapping, %3D scanning,
%Augmented Reality (AR) and/or Mixed Reality (MR) systems.
%Augmented Reality 
AR
and/or 
%Mixed Reality 
MR
systems~\cite{Bimber:2005:SAR:1088894}.
%pcompution
%to efficiently present information to users by projecting various 
%images onto a scene or object surfaces.
%In these applications, since the depth of field found in common projectors is 
%usually narrow, defocus blur inevitably occurs in practice, which is one of the 
%critical issues for practical usage of AR and MR systems. Recently, laser 
%projectors which theoretically have no defocus blur have been developed and made 
%commercially available. By using such defocus-free projectors, new applications 
%are expected to be developed for various AR/MR scenarios.
 %, research areas and practical situations.
%For instance, 
%Since such defocus free projectors can project patterns at any distance, one may 
%consider that 
%Apart from 3D scanning system, 
%In projection mapping and AR/MR, the 
In those systems,
the precalculated scene 
geometry is used to project an appropriately warped image to be mapped on the 
scene as an artificial texture, with impressive visual results. However, 
it is difficult 
to layout 
with moving objects or dynamic scenes in general.
This is because the projected pattern is the same along each ray, hence what is viewed is 
spatially invariant up to a projective transformation. Conversely, the potential 
for practical applications could be significantly broadened if different 
patterns can be projected at different depths simultaneously;
%For instance, by considering projection mapping using multiple semi-transparent 
%screens, different movies can be projected on each screen. 
%If different depth layers from 
%a scene are projected on parallel screens placed at different depths for example, this could effectively 
%increase the users' three-dimensional perception; 
such volume displays are now 
intensely 
investigated~\cite{barnum2010multi,jurik2011prototyping,nagano2013autostereoscopic,hirsch2014compressive}. 
However, those techniques are still under research and are still not suitable for practical systems.
Recently, a simpler system consisting of just two projectors which can 
simultaneously project independent images on screens 
at different depths has been proposed~\cite{Scarzanella:psivt15}. The system demonstrated the ability to project simultaneous movies on multiple semi-transparent screens at different depths with no particular setup requirements, which is a promising research avenue to explore.

%Similarly, in a scene exhibiting dynamic, local 
%geometry changes, different patterns could be visualised according to the 
%changing scene depth without the need for explicit 3D reconstruction and/or 
%change of projected images. A system able to project different patterns at 
%different predefined depths can also be used as a non-contact three-dimensional 
%measurement device, which can be used for manufacturing purposes or to aid 
%visual assessment of distances to avoid, for example, vehicle collisions. 
%with high precision.

\begin{figure}[tb]
\vspace{-0.5cm}
\centering
        \includegraphics[width=0.88\textwidth]{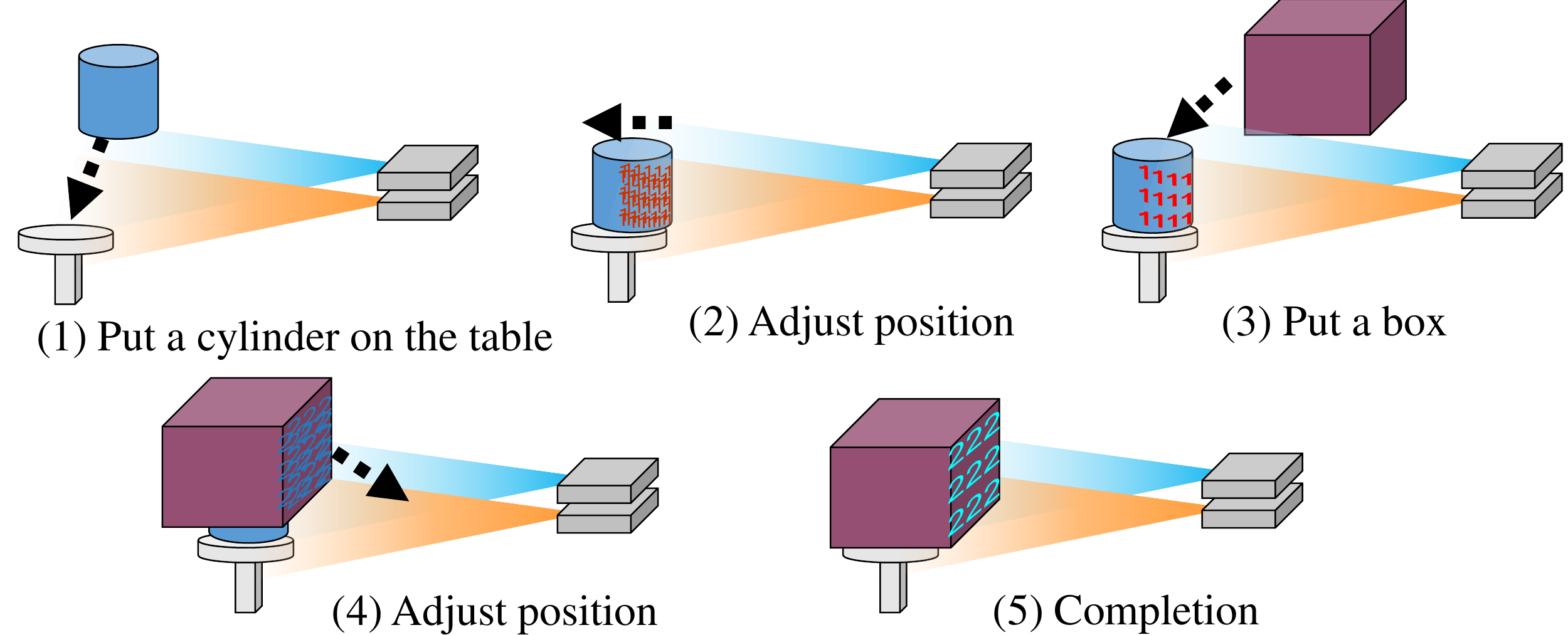}
\vspace{-0.2cm}
	\caption{Basic scheme to align two objects at the right position by 
    visual feedback using two projectors. Note that projected patterns are static.}
	\label{fig:app}
	\vspace{-5mm}
\end{figure}

In this paper, we significantly extend the system~\cite{Scarzanella:psivt15} by 
%\fnoteII{
1) removing the planarity assumption: in this work, we are able to project depth-dependent patterns on complex surfaces with arbitrary geometry, 2) dynamic range expansion by adding a constraint in pattern optimization, and 3) introducing epipolar constraint to keep the problem realistic size.
%}
With our extension, the potential 
application is significantly broadened. For instance, we can use the system for 
object placement or assembling purposes as shown in Fig.~\ref{fig:app}: since the pattern is sharply visible only at predefined depths,  
human or robotic workers can precisely align complex parts using only
qualitative visual information, without distance scanners. While the visibility of the pattern depends on its 3D position and orientation in space, translational alignment can also be achieved by including markers in the pattern. The new system also allows dynamic projection mapping on complex and semi-transparent objects.
%
%We further contribute by describing a practical optimization scheme to break away from the limitations found in the literature, where the colour dynamic range of the pattern projected is severely limited~\cite{Scarzanella:psivt15,nakamura2011emphasizing}. Finally, in our new formulation we introduce epipolar constraints to decompose the global pattern generation process into multiple local ones, which allows to exploit parallelism and to avoid the risk of getting stuck in local optimization minima.
In the experiments, we show the effectiveness of the proposed technique, which is also demonstrated on multiple dynamic 3D solid objects.

\section{Related work}
\label{sec:related}

%\fnoteII{
%[Furukawa]
%Now, this section is the same as the PSIVT paper. The expressions should be revised.
%It would be better if some related works can be added. }
%\knote{[Kawasaki] Add our PSIVTwsVG paper.}

%\knote{[Kawasaki] Add three references.}

Applications of multiple video projectors have a long history in VR systems
such as wide or surround screen projections like CAVE \cite{cruz1993surround}.
These types of
% surround-screen 
projectors needs precise calibration between projectors,
\ie,  %, 
%both geometric and photometric. 
the geometrical calibration to establish the correspondences 
%between the pixels on the screen and each projector's pixels, and to calculate the overlap between projectors through the calculated planar homographies. 
between the projected images and  the screens,
%, and to calculate the overlaps between the projectors. 
and the photometric calibration
% is also necessary 
to compensate for nonlinearities between the projected intensities and the pixel values. 
%Moreover, depending on the relative distance and angle between projectors, other scene-dependent photometric attenuation effects must be taken into account. 
To this end, 
automated calibration techniques based on projector-camera feedback 
systems were developed~\cite{raskar1998seamless,yang2001pixelflex}.
Since some of the screens considered was curved, 
some of these works inevitably dealt with non-planar screens. 
In other works, multiple projectors were tiled together 
%used 
for improving the 
resolutions of the projected images~\cite{chen2000automatic,Schikore2000CGA}. 
Godin \etal, inspired by human vision system, 
proposed a dual-resolution display where the central part of the projection is projected in 
high-resolution, while the peripheral area is projected in low-resolution
%, 
%mirroring the structure of foveal vision
~\cite{godin2006high}.
%Further applications of 
Other works on multi-projector systems focused on increasing the depth-of-field,  
since this is normally narrow and can cause defocusing issues on non-planar screens.
Bimber and Emmerling~\cite{Bimber2006} proposed to widen the depth-of-field
by using multiple projectors with different focal planes.
Nagase \etal ~\cite{nagase2011dynamic}
used an array of mirrors,
which is equivalent to multiple projectors with different focal planes,
for correcting defocus, occlusion and stretching artifacts. 
Levoy \etal~also used an array of mirrors and a projector~\cite{levoy2004synthetic}. The array of mirrors was used to avoid occlusions from objects placed in front of the screen. 
Each of the aforementioned works is intended to project a single
image onto a single display surface, which may or may not be planar.
Conversely, 
the proposed method
projects multiple, independent images onto surfaces placed at different depths, which may have a complex non-planar geometry. 

Multiple projectors are also used for 
applications in light-field displays~\cite{jurik2011prototyping,nagano2013autostereoscopic}.
For these applications, in order to create the large number of rays needed for the light field,
each ray is projected separately for a specific viewpoint and is not intended 
to be mixed with other rays.  
This is in contrast with our proposed method, 
where the multiple independent images are created at the intended depths and on the intended surfaces exactly by leveraging the mixing properties of rays from the projectors.

%Overall, although the act of combining pixel values cooperatively from multiple 
%projectors that share a common object point could be optimized for numerous 
%tasks, algorithms and applications have not been well explored yet in the 
%community. 

\begin{figure}[t]
\vspace{-0.3cm}
\centering
	\includegraphics[width=1\textwidth]{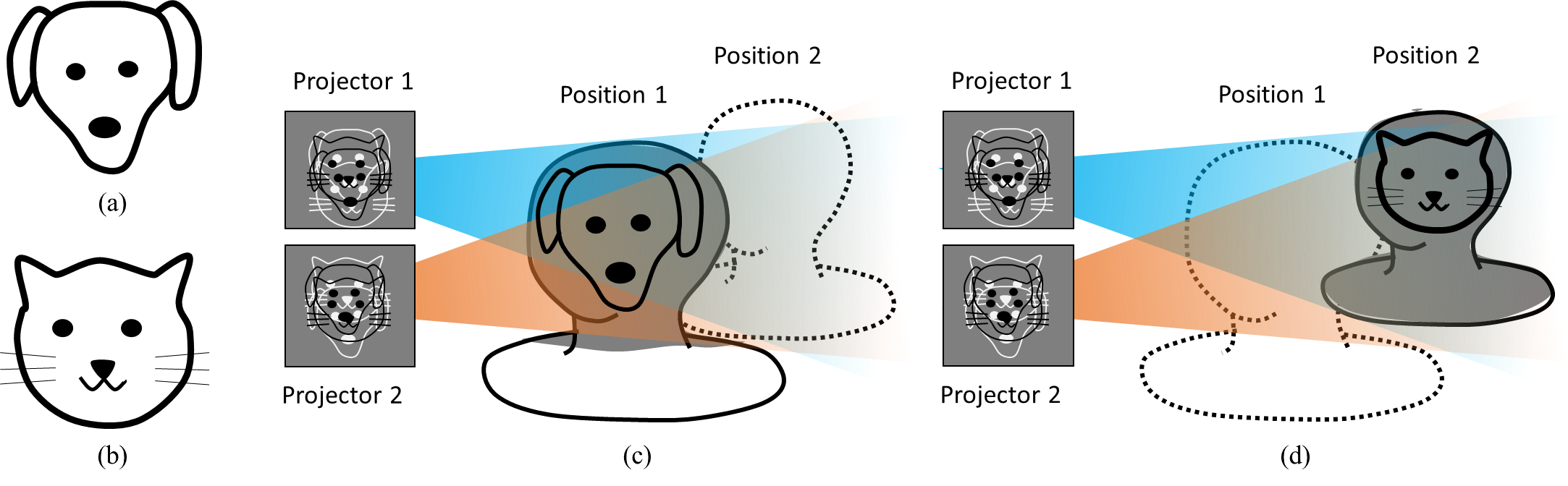}
	\caption{Projection of virtual masks on dynamic subjects.
	To project the masks in images (a) and (b),
	specially generated images should be projected.
	When the subject is at position 1, only mask (a) is shown, whereas (b) is suppressed, as shown in (c). The situation is inverted in (d).}
	\label{fig:basic}
\vspace{-0.3cm}
\end{figure}

A few works exist that have explored the concept of 
``depth-dependent projection''; 
%There are three systems that propose techniques for highlighting 3D structure 
%according to its depth using multiple projectors. 
Kagami 
\cite{Kagami2010}
projected
Moire patterns to visually show the depth of a scene,
similarly to active-stereo methods for range finding. 
%structure and depth is highlighted by projecting interfering Moire patterns or 
%complementary colours. 
Nakamura \etal
~\cite{nakamura2011emphasizing}
used a linear algebra formulation to highlight predetermined volume 
sections 
with specific colors. 
The technique assumes the volume is discretized into multiple parallel planes and is not able to produce detailed patterns or images on non planar surfaces. Moreover, similarly to the work below, the underlying mathematical formulation suffers from a limited dynamic range. 
%Conversely, we propose a novel 
%application for the display of detailed images at distinct locations in space by 
%actively exploiting interference patterns from multiple projectors. Furthermore, 
%in Section \ref{sec:detail} we highlight the differences in the formulation, 
%which allows us to exploit the sparse structure of the problem and to solve it 
%very efficiently despite very large matrix sizes.
Recently, Visentini-Scarzanella \etal~proposed a method to display detailed images on distinct planes in space by 
actively exploiting interference patterns from multiple projectors~\cite{Scarzanella:psivt15}.
In their experiments, two independent videos are simultaneously streamed on two 
screens using semi-transparent material. However, the matrix factorisation used is similar to \cite{nakamura2011emphasizing}, so the method suffers from a limited dynamic range. Moreover, the purpose is to project images on planar screens, and thus, applications are limited. 
These limitations are removed in our proposed method, were the patterns can be projected onto arbitrary shapes, and our novel optimization procedure addresses the issues with dynamic range.

%\section{System Overview} % \knote{Kawasaki\&Marco}}
\section{Algorithm Overview} % \knote{Kawasaki\&Marco}}
\label{sec:overview}

%\fnoteII{
%[Marco]
%The explanation using planes should be avoided. 
%Thus, the images should be changed so that the targets are curved surfaces.
%The gray-code projection for acquiring the mapping between the projector and the surface
%should be also mentioned here. }

%\begin{figure}[tb]
%\vspace{-0.3cm}
%\centering
%	\begin{subfigure}{0.4\linewidth}
%		\includegraphics[width=1\textwidth]{proj1-eps-converted-to.pdf}
%		\caption{}
%		\label{fig:basica}
%	\end{subfigure}
%		\begin{subfigure}{0.4\linewidth}
%		\includegraphics[width=1\textwidth]{proj2-eps-converted-to.pdf}
%		\caption{}
%		\label{fig:basicb}
%	\end{subfigure}\\
%		\begin{subfigure}{0.4\linewidth}
%		\includegraphics[width=1\textwidth]{proj3-eps-converted-to.pdf}
%		\caption{}
%		\label{fig:basicc}
%	\end{subfigure}
%		\begin{subfigure}{0.4\linewidth}
%		\includegraphics[width=1\textwidth]{proj4-eps-converted-to.pdf}
%		\caption{}
%		\label{fig:basicd}
%	\end{subfigure}
%\knote{\\This figure should be removed or totally modified to simply show the concept.}
%\vspace{-0.3cm}
%	\caption{Basic scheme to create patterns for two depths with two projectors.}
%	\label{fig:basic}
%\vspace{-0.3cm}
%\end{figure}

We provide an outline of the algorithm using the example of projecting virtual face masks on dynamic subjects as shown in Fig. \ref{fig:basic}.
In Figs. \ref{fig:basic}a and \ref{fig:basic}b two different virtual masks are shown, 
which are to be projected on the subject's face when this changes position from Position 1 to Position 2 as shown in Figs. \ref{fig:basic}c and \ref{fig:basic}d respectively. 
The fundamental research question is what images should be generated for Projector 1 and Projector 2 so that the virtual masks would recombine into the patterns in Figs. \ref{fig:basic}a and \ref{fig:basic}b at the desired position: at position 1, only mask of Fig. \ref{fig:basic}a should appear, with no traces of mask of Fig. \ref{fig:basic}b. 
%\fnoteII{
It should be noted that similar problem was raised for multiple LCDs and efficiently solved by \cite{Wetzstein:2012:TensorDisplays}.
%}

%Conversely, at position 2 only mask of Fig. \ref{fig:basic}b should appear projected on the subject, as if an active system based on 3D tracking were present. 
%Significantly, if the patterns were to be projected on semitransparent screens at the correct locations, both patterns would simultaneously appear.

Visentini-Scarzanella \etal~realised a similar system \cite{Scarzanella:psivt15}, 
but planar screens were assumed. Moreover, significant reduction in the dynamic range was observed due to their formulation.
Because of the sensitivity of \cite{Scarzanella:psivt15} to the exact screen placement, it is not possible to directly apply the method to project onto objects with complex geometry by simply considering a piecewise planar approximation.

\begin{figure}[t]
\vspace{-0.3cm}
	\centering
	\begin{subfigure}{0.46\textwidth}
%	\includegraphics[width=6cm]{figures/configuration.jpg}
%\vspace{5cm}
%	\includegraphics[width=1\textwidth]{system_image.JPG}
	\includegraphics[width=1\textwidth]{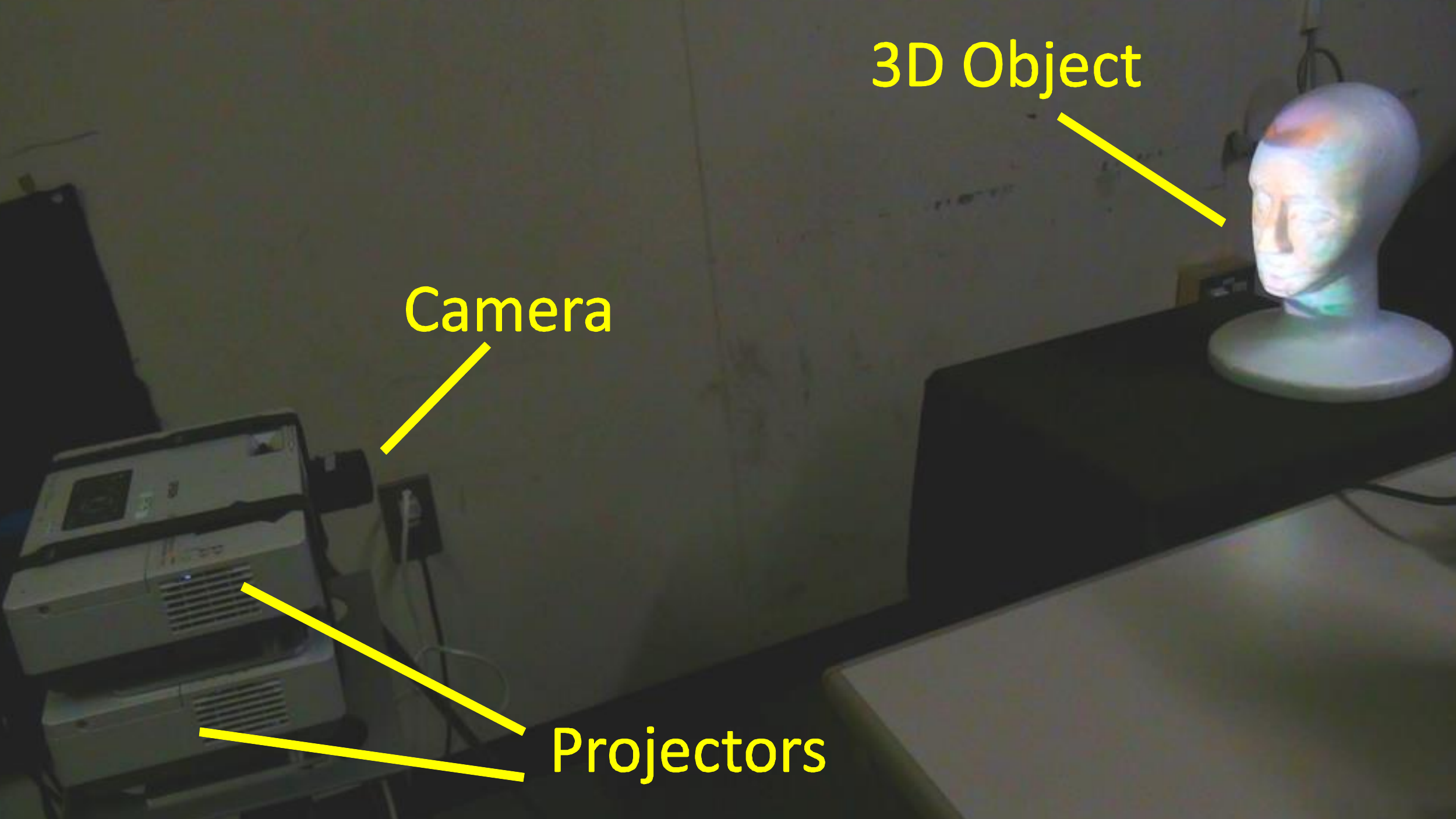}
%\knote{[Hirukawa] Change image to box version.}
	%\includegraphics[width=3.419cm]{figures/conf02.jpg}\\
%\knote{from Masuyama slide.}
\caption{}
\label{fig:config}
\end{subfigure}
\begin{subfigure}{0.53\textwidth}
\includegraphics[width=1\textwidth]{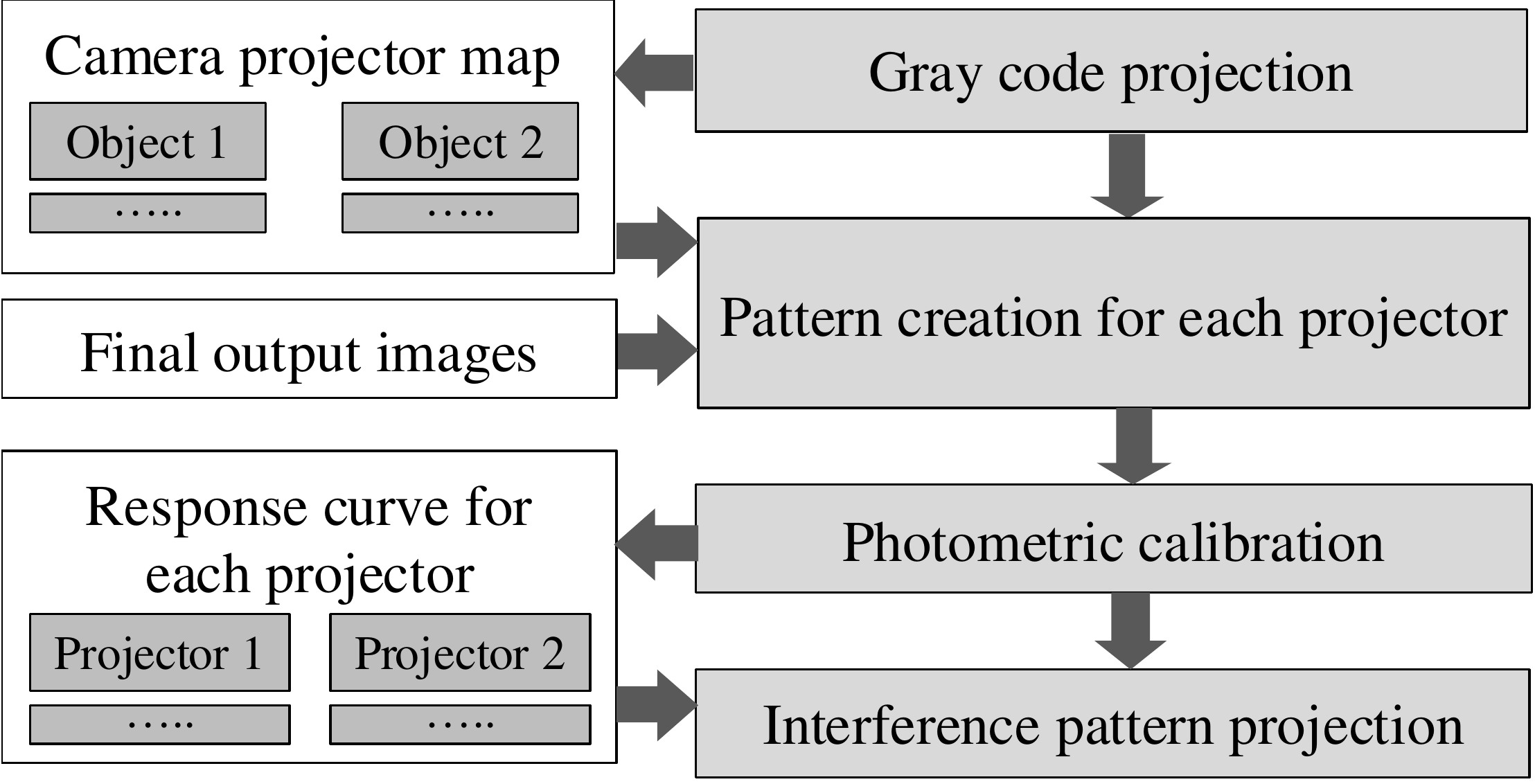}
%\knote{[Kawasaki] Change image to four depth (box) version.}
\caption{}
\label{fig:diag}
\end{subfigure}
\vspace{-0.3cm}
\caption{(a) Configuration of our system with two projectors. (b) Overview of the algorithm.}
%	\vspace{-4mm}
\end{figure}

In order for the desired images to appear at the desired locations, three tasks are necessary. 
First, the mapping between points on the object surfaces and the pixels in the projector images
are obtained (\ie, geometrical calibration).
Then, given the mappings, generation of the projection images can be cast as a constrained optimization problem. 
%Since each ray from each projector intersects multiple rays from the other projector, altering a single pixel value to change one element on the object surface will require adjusting multiple values from related pixels. 
In~\cite{Scarzanella:psivt15}, this was solved globally with a sparse matrix solver that distributes the error throughout the images. In this paper, we propose an efficient optimization method to solve the problem locally only for related rays, allowing to impose additional constraints on the solution resulting in improved dynamic range, as well as to allow parallelisation.
Finally, the generated patterns have to be post-processed,
according to the photometric characteristics obtained in the process of photometric calibration,
to correct non-linearity of the projector. 
The main phases of the 
algorithm are shown in Fig.~\ref{fig:diag}.
%of the scene and the projectors, to compensate for the relative object distances as well as the intensity response curve of each projector.

%One may consider whether the process always converges to 
%create valid patterns. In this paper, it is revealed that the problem cannot be 
%solved exactly because of the finite field of view of the pattern of projector, 
%however, at the same time, it is shown that close approximations can be created 
%by distributing the approximation error over the whole projected pattern image.
%we propose the method which can make meaningful result by 
%distributing the error over the projected pattern.

%\knote{[Kawasaki] Add more planes, at least four for 2 boxes.}

The system consists of two standard LCD projectors stacked vertically as shown in Fig.~\ref{fig:config}
 with 3D objects for the projection target that can be placed at arbitrary positions.

\begin{figure}[t]
%\vspace{-0.3cm}
	\begin{subfigure}{1\textwidth}
	\centering
		\begin{subfigure}{0.12\textwidth}
			\includegraphics[clip=true,width=1\textwidth]{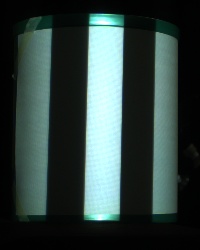}
			\caption{}
			\label{fig:homob}
		\end{subfigure}
		\begin{subfigure}{0.12\textwidth}
			\includegraphics[clip=true,width=1\textwidth]{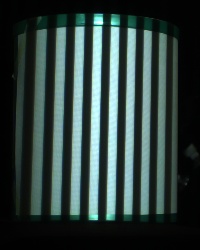}
			\caption{}
			\label{fig:homoc}
		\end{subfigure}
		\begin{subfigure}{0.12\textwidth}
			\includegraphics[clip=true,width=1\textwidth]{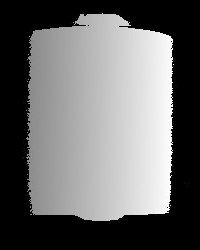}
			\caption{}
			\label{fig:homod}
		\end{subfigure}
		\begin{subfigure}{0.12\textwidth}
			\includegraphics[clip=true,width=1\textwidth]{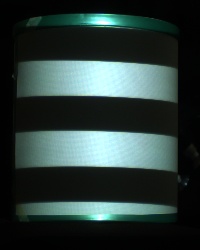}
			\caption{}
			\label{fig:homoe}
		\end{subfigure}
		\begin{subfigure}{0.12\textwidth}
			\includegraphics[clip=true,width=1\textwidth]{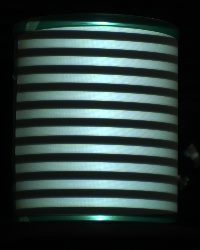}
			\caption{}
			\label{fig:homof}
		\end{subfigure}
		\begin{subfigure}{0.12\textwidth}
			\includegraphics[clip=true,width=1\textwidth]{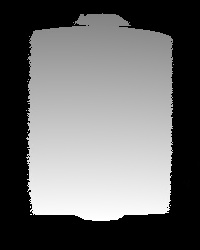}
			\caption{}
			\label{fig:homog}
		\end{subfigure}
		\end{subfigure}\\
%	\vspace{0.3cm}
		\centering

	\begin{subfigure}{0.7\textwidth}
		\centering
%\vspace{-0.5cm}
%	\includegraphics[width=0.7\textwidth]{homography01.pdf}
%	\includegraphics[width=0.5\textwidth]{homography01.eps}
%	\includegraphics[width=1.0\textwidth]{graycode_cam.eps}
	\includegraphics[width=1.0\textwidth]{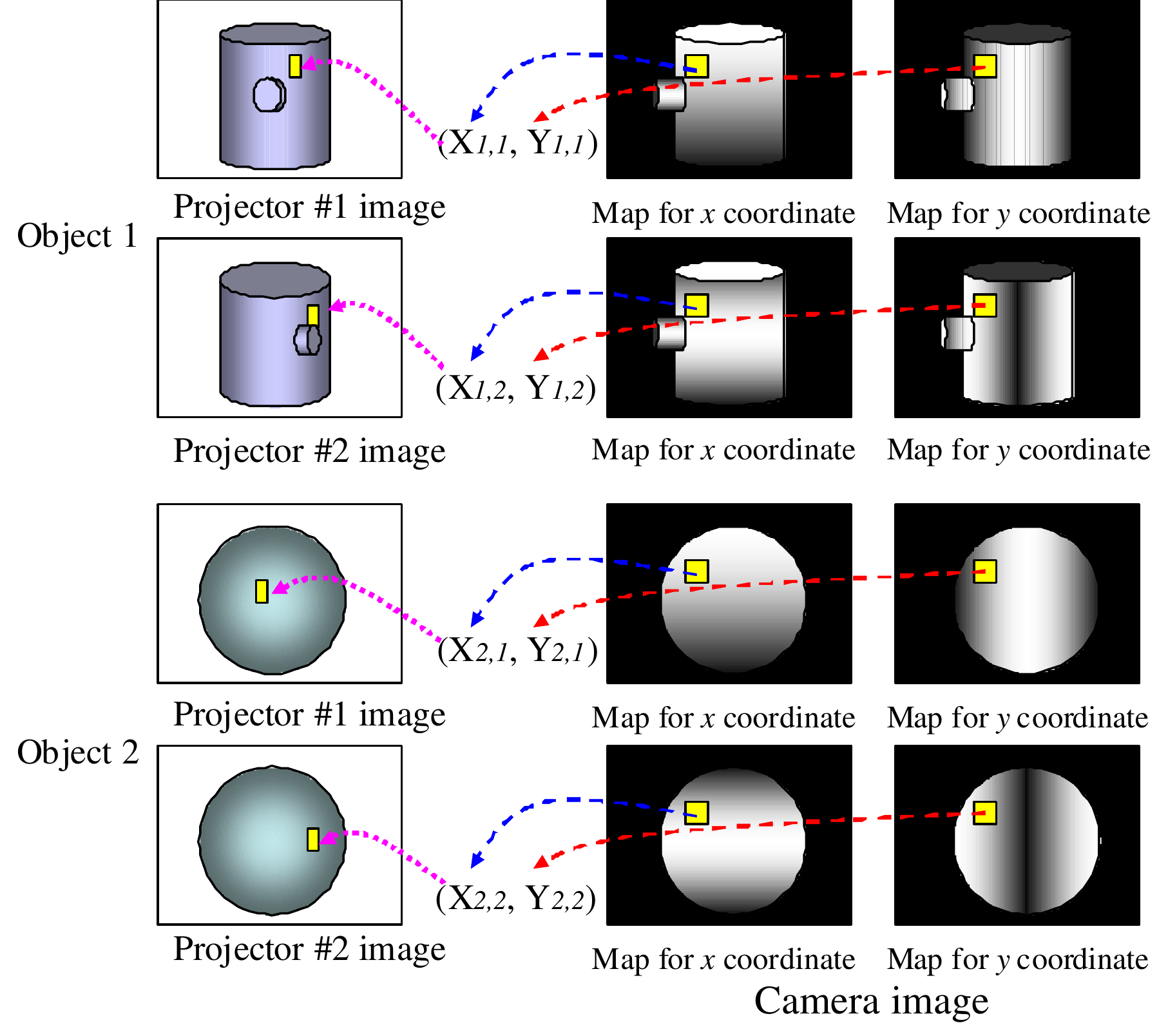}
\vspace{-0.8cm}
	\caption{}
	\label{fig:homoa}
%\knote{[Hirukawa] Change image to a box version.}
	\end{subfigure}

%	\begin{subfigure}{0.7\textwidth}
%		\centering
%%\vspace{-0.5cm}
%%	\includegraphics[width=1.0\textwidth]{graycode_proj.eps}
%	\includegraphics[width=1.0\textwidth]{graycode_proj-eps-converted-to.pdf}
%\vspace{-0.8cm}
%	\caption{}
%	\label{fig:homoa}
%%\knote{[Hirukawa] Change image to a box version.}
%	\end{subfigure}

\vspace{-0.3cm}
%	\caption{(a) Calibration board for plane/camera homography. (b),(c) Composite images with calibration board and projected checkerboard pattern for camera/projector homography calculation from projector 1 and 2 respectively. (d) Required homographies.}
\caption{(a),(b) An object of 3D screen projected by Gray code patterns,
(c) projector coordinates observed in the camera coordinates for x direction,
(d),(e),(f) for y direction, and
(g) mapping between projector image coordinates (the left column) and 
the camera image obtained from Gray code (the two right columns).
% attached with checker patterns. (b),(c) Composite 
%images with attached pattern and projected pattern using estimated  
%    homographies at different depths. (d) 
%Required homographies.
}
	\label{fig:graycode}
\vspace{-0.4cm}
\end{figure}

\section{Depth-dependent simultaneous image projections}
\label{sec:detail}

%\subsection{Geometric calibration} % \knote{Kawasaki\&Marco}}
\subsection{System calibration} % \knote{Kawasaki\&Marco}}
\label{sec:calib}
\label{sec:calib_response}

%\fnoteII{
%[Kawasaki]
%The calibration step is replaced by explanation of the Gray Code projection. 
%}

%\subsubsection{Geometric calibration}
%\knote{[Kawasaki] Add more planes, at least four as for the box case.}5

To achieve 
%our goal of 
depth-dependent projections onto arbitrarily shaped objects, it is necessary to estimate the pixel correspondences between  
projector images and the object surfaces. Contrarily to the simple planar case in~\cite{Scarzanella:psivt15} where the mapping can be calculated as a homography using only four corresponding points, 
%\fnoteII{
we need to obtain correspondences between the projector pixels and the points on complex-shaped 3D surfaces.
%}
%we introduce an additional camera and projection of 
%and a special projection pattern encoding the positional information of the projector pixel coordinates. 
%\fnoteII{
To this end, we use the Gray 
code pattern projection~\cite{sato1985three}.
%which is one of the most popular patterns 
%to efficiently encode the positional 
%information into multiple patterns 
%as shown in Figs.\ref{fig:graycode}a, 
%\ref{fig:graycode}b, \ref{fig:graycode}d and \ref{fig:graycode}e.
%}
%Final decoded values are shown in Figs.\ref{fig:graycode}c and 
%\ref{fig:graycode}f where pixel value represents the projector coordinate for x 
%and y direction, respectively.

The actual process is as follows. First, an additional camera is placed in the scene. Then, the Gray 
code patterns are projected onto the object and the projections are synchronously captured by the camera
as shown in Figs.\ref{fig:graycode}a, 
\ref{fig:graycode}b, \ref{fig:graycode}d and \ref{fig:graycode}e..
By decoding the pattern from the captured image sequences, the correspondences between 
projected patterns $\textbf{P}_j$ and the camera image coordinates $\textbf{C}$ are obtained.
The decoded values are shown in Figs.\ref{fig:graycode}c and 
\ref{fig:graycode}f where pixel value represents the projector coordinate for x 
and y direction, respectively.
The decoded results are represented by a map 
$f_j : \textbf{C}\rightarrow\textbf{P}_j$, which consists of projector coordinate 
values embedded at each pixel of the camera image as shown in 
Figs.\ref{fig:graycode}c, \ref{fig:graycode}f. % and \ref{fig:graycode}e right column.
An inverse map is also obtained
$f^{-1}_j : \textbf{P}_j\rightarrow\textbf{C}$
%as shown in Fig.\ref{fig:graycode}(c) 
to efficiently conduct the 
extraction of corresponding points along
the epipolar lines as described in the following section. Note that because of mismatch between the resolutions of the image and the projection on the object, many correspondences 
are inevitably dropped from the maps, and thus, the final results are degraded. 
Such artifacts are mostly solved by preparing a high 
resolution map if projectors and camera can be placed close to each other. 
%These issues are addressed in the experimental section.
The remaining holes are in the order of a few pixels, and are removed with a simple hole filling algorithm.

\begin{figure}[t]
	\centering
	\begin{subfigure}{0.40\textwidth}
		\includegraphics[width=1\textwidth]{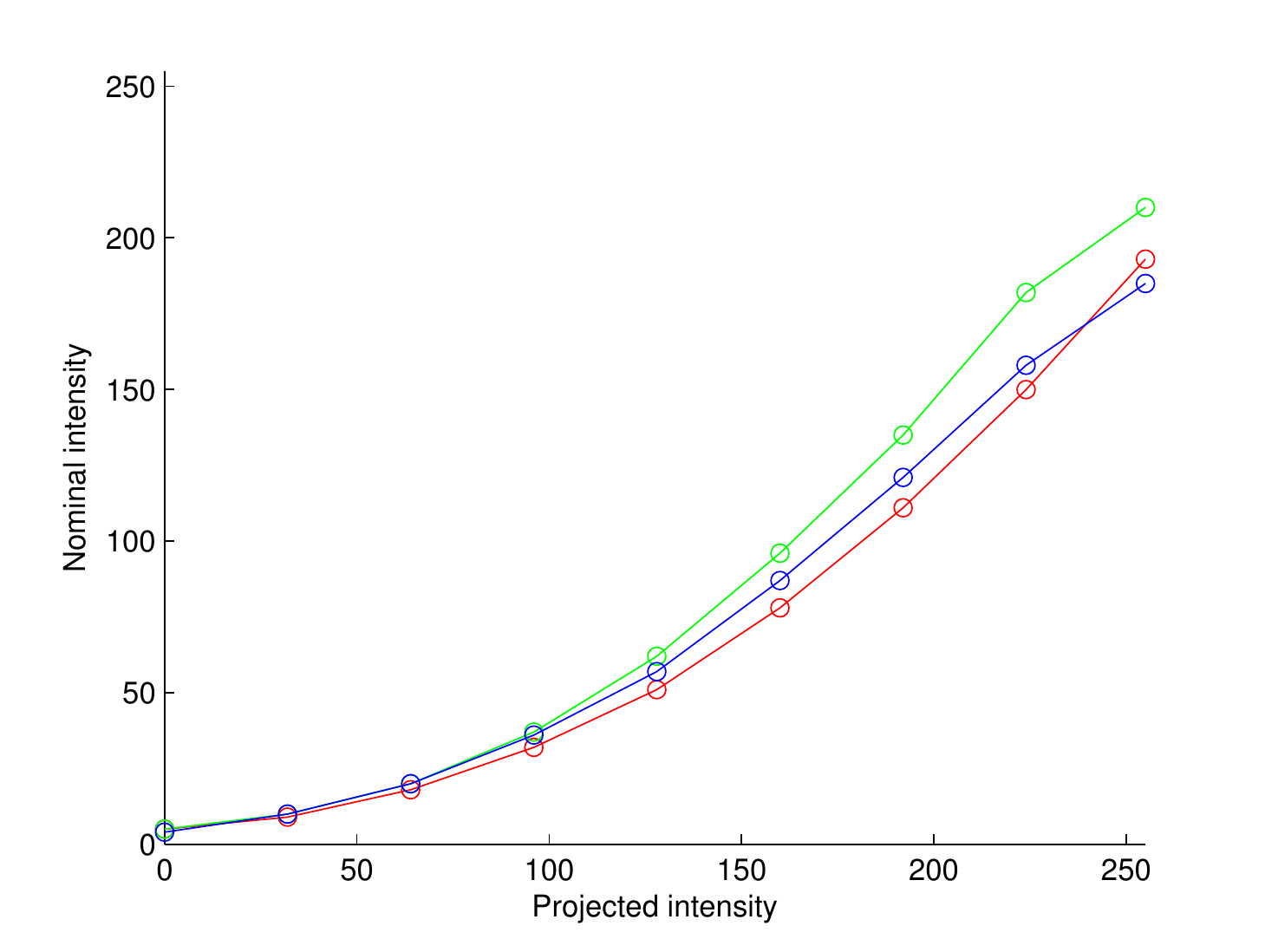}
	\caption{}
	\label{fig:response1}
	\end{subfigure}
	\begin{subfigure}{0.40\textwidth}
		\includegraphics[width=1\textwidth]{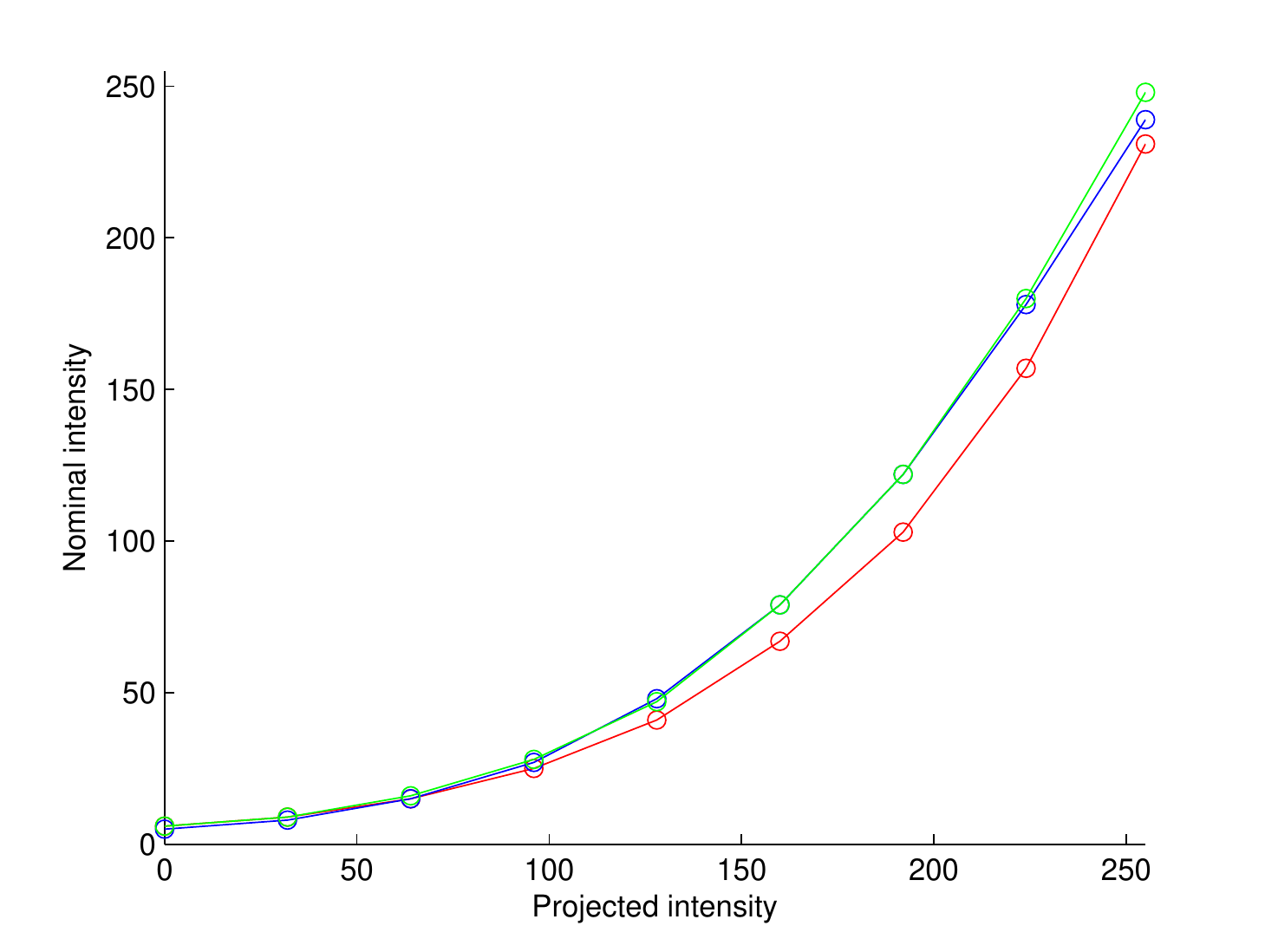}
	\caption{}
	\label{fig:response2}
	\end{subfigure}
	\centering
	\begin{subfigure}{0.125\textwidth}
		\includegraphics[width=1\textwidth]{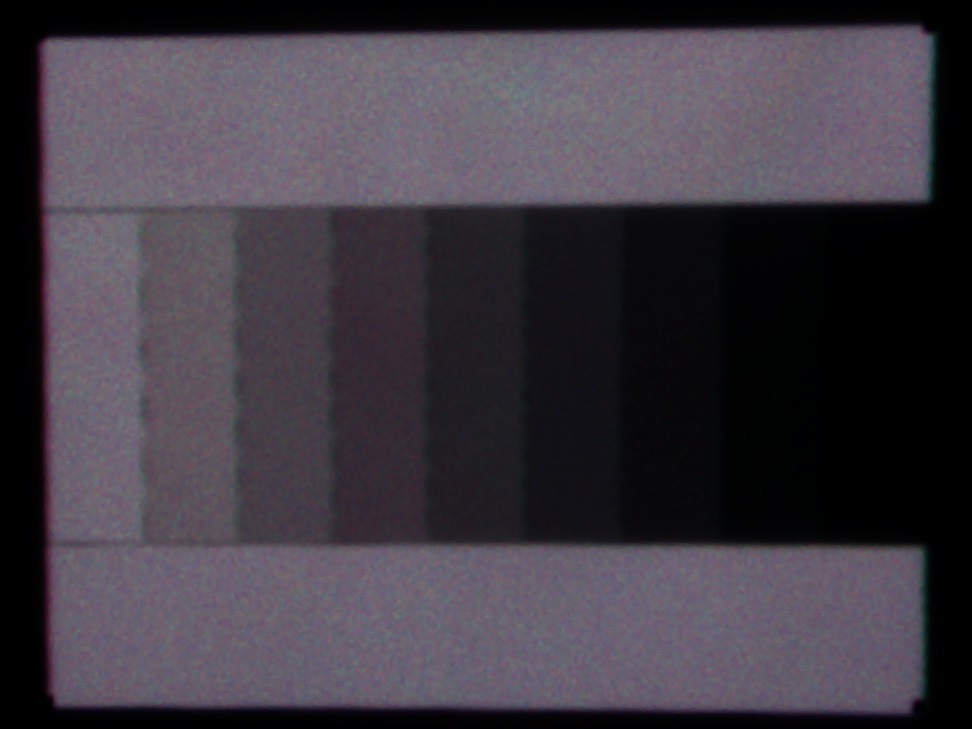}
		\includegraphics[width=1\textwidth]{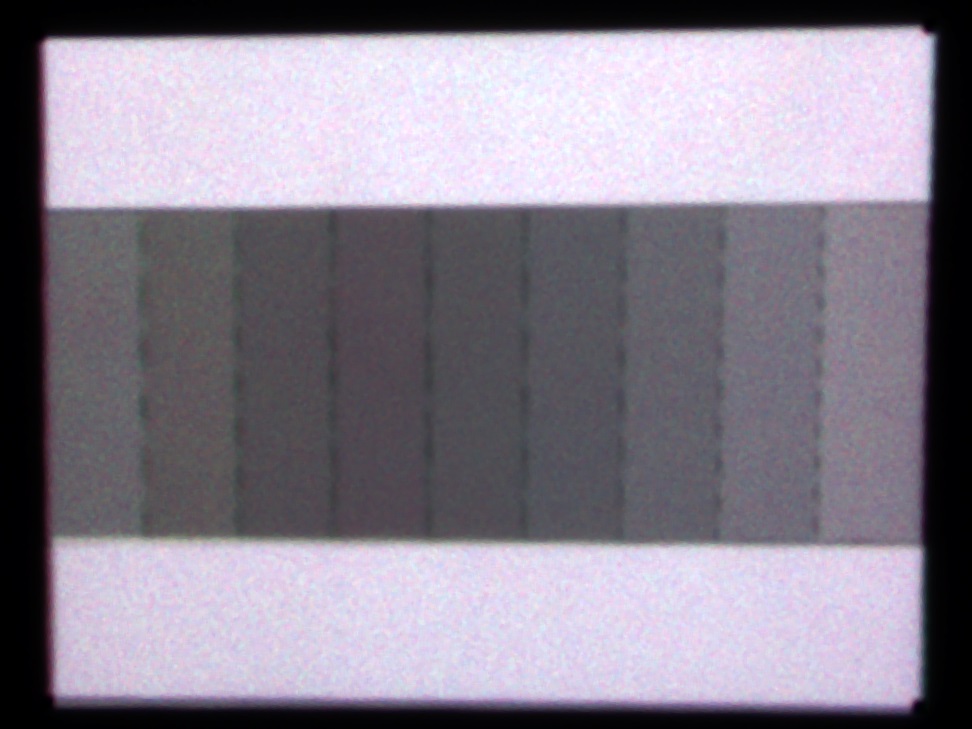}
		\includegraphics[width=1\textwidth]{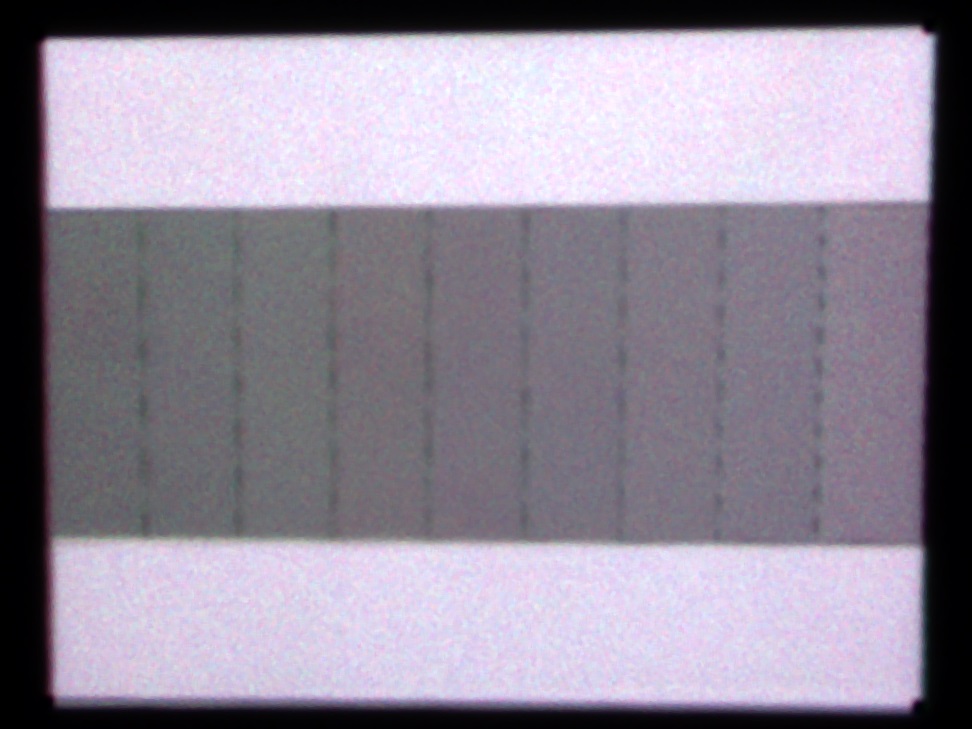}
	\caption{}
	\label{fig:cap_respb}
	\end{subfigure}
%	\begin{subfigure}{0.22\textwidth}
%		\includegraphics[width=1\textwidth]{photocalib1-eps-converted-to.pdf}
%	\caption{}
%	\label{fig:response1}
%	\end{subfigure}
%	\begin{subfigure}{0.22\textwidth}
%		\includegraphics[width=1\textwidth]{photocalib2-eps-converted-to.pdf}
%	\caption{}
%	\label{fig:response2}
%	\end{subfigure}\\
%	\centering
%	\begin{subfigure}{0.125\textwidth}
%	\includegraphics[width=1\textwidth]{proj1resp.jpg}
%	\caption{}
%	\label{fig:cap_respb}
%	\end{subfigure}
%	\begin{subfigure}{0.125\textwidth}
%	\includegraphics[width=1\textwidth]{proj12merge.jpg}
%	\caption{}
%	\label{fig:cap_respc}
%	\end{subfigure}
%	\begin{subfigure}{0.125\textwidth}
%	\includegraphics[width=1\textwidth]{proj12merge_compensated.jpg}
%	\caption{}
%	\label{fig:cap_respd}
%	\end{subfigure}
	\vspace{-2mm}
	\caption{(a),(b) Intensity response curves for projectors 1 and 2 
respectively. (c) (top)Projected calibration pattern, and (middle) calibration pattern 
superimposed with its own mirrored version, before and (bottom) after colour 
compensation.}	\label{fig:cap_resp}
	\vspace{-3mm}
\end{figure}
In the process of estimating the projected image,
% which will be described later, 
we assume a linear relationship between the nominal intensity and the actual projected intensity. 
%Then, it is necessary to ensure that there is a linear relationship between the nominal intensity and the actual projected intensity.  
%Indeed, 
%experimentally it was found that whenever 
%%this stage 
%the photometric calibration
%was omitted, large errors 
%were visible in the recombined images.
%, even when both projectors were identical as shown in Fig. \ref{fig:comparison}.
%To this end, 
To achieve this, 
%For photometric calibration, 
a linearly 
increasing gray scale pattern in the $[0,255]$ intensity range (Fig.~\ref{fig:cap_respb}(top))
%. This 
is captured by a camera with a known linear response.
% and the median value for each of the RGB 
%channels is taken for each intensity bar. 
The recorded values are plotted against their nominal intensity, as shown 
 in Fig.~\ref{fig:response1}, 
\ref{fig:response2}, which are then fitted to a $f(x)= ax^b + c$ model, 
where $x$ is the intensity value. The function is then inverted and kept for 
compensating the generated pattern prior projection.
The calibration pattern superimposed with its own mirrored version
should be constant intensity. By intensity correction, 
this constraint is shown to be fulfilled(Fig. \ref{fig:cap_respb}(middle and bottom)).

\subsection{A simple linear algebra-based pattern generation method}%Problem definition}
%\subsection{Interference pattern generation via epipolar optimisation}%Problem definition}
\label{sec:make_pat}

%\knote{[Furukawa] Explain defocus effects. Is this section a right place?}

\begin{figure}[t]
	\centering
	\includegraphics[width=0.8\textwidth]{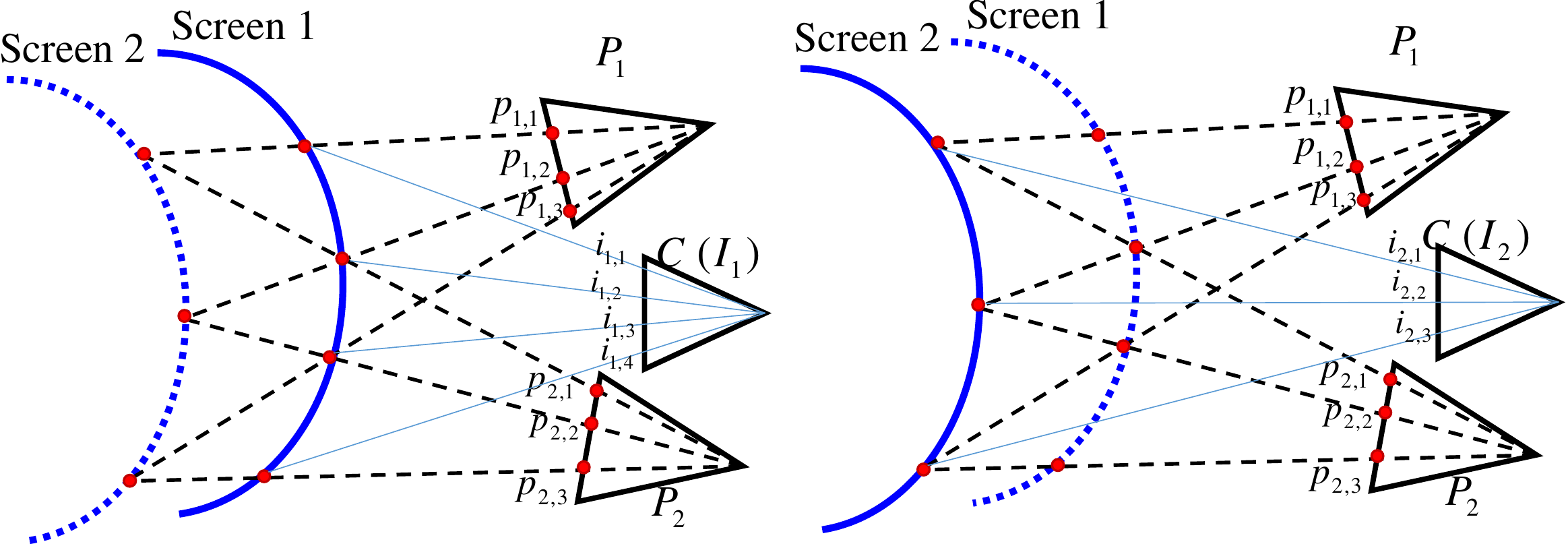}
	\caption{Variables of linear constraints.}
	\label{fig:variables}
%	\vspace{-2mm}
\end{figure}

%\knote{[Furukawa] Add planes, at least four.}

%\fnoteII{
%[Marco] Revise the formulations. [But basic formulation does not change, doesn't it?]
%}

We model the problem by first extending the formulation in~\cite{Scarzanella:psivt15} to the case of non-planar surfaces. The variables involved are shown in Fig.~\ref{fig:variables}. While 
for clarity we illustrate the process in the case of two projectors and two 
different images 
projected onto two different objects,
%placed at two depth levels, 
the system can be extended to a 
higher number of projectors and 
objects. 
%depth planes. 

The projected patterns from $J$ projectors are denoted as 
${\bf P}_j$ where $j \in \{1,\cdots,J\}$
($J=2$ in case of Fig.~\ref{fig:variables}),
and $K$ images 
%to be 
%shown at the two different depths are 
projected on $K$ 3D objects are
depicted as 
${\bf I}_k$ where $k \in \{1,\cdots,K\}$
($K=2$ in case of Fig.~\ref{fig:variables}).
Let pixels on ${\bf P}_j$ be expressed as $p_{j,1},p_{j,2},\cdots,p_{j,m},\cdots,p_{j,M}$
and let pixels on ${\bf I}_k$ be $i_{k,1},i_{k,2},\cdots,i_{k,n},\cdots,i_{k,N}$.

We also need to provide the mapping from 
the desired input images ${\bf I}_k$ 
to the camera coordinates
${\bf C}$, which is
$g_k : {\bf I}_k \rightarrow {\bf C}$.
Practically, 
in this paper, 
we use identity map for $g_k$,
which means we use simple ``projection mapping''
as shown in 
case of Fig.~\ref{fig:variables}, 
where 
the coordinates of ${\bf I}_1$ and 
${\bf I}_2$ is the same as camera image of ${\bf C}$.
%If we need more complex mapping designed for the shape of the 
%3D screens,
%aligning the 3D positions and the CAD model of the 3D screens
%is required. 

In the calibration step, 
the mappings $f_j$ have been obtained. 
Using $f_j$ and $g_k$, 
we can map between the pixels of the projected image
${\bf P}_j$ and the 
desired input image ${\bf I}_k$
with $( f_j \circ g_k ) : {\bf I}_k \rightarrow {\bf P}_j$
and $(f_j \circ g_k )^{-1}: {\bf P}_j \rightarrow {\bf I}_k$.

%Here, we assume that each of the $K$ objects can be
%modeled as polygons,
%%and the mapping from the projected image to a polygon
%%are known as 
%and the mapping from a projected image $P_j$ to 
%each of the polygons are modeled as a homography
%%The image projection from $P_j$ to 
%%$I_k$ can be modeled as a homography 
%with the parameters estimated during calibration. 
%%To model the image projection, 
%The image projection from $P_j$ to 
%$I_k$ can be modeled as multiple homographies.

%Using these parameters, 
From these assumptions, 
we can define an inverse projection mapping $q$, 
where, 
if $i_{k,n}$ is illuminated by $p_{j,m}$
(\ie, pixels of $i_{k,n}$ and  $p_{j,m}$ are mapped with $( f_j \circ g_k )$),
$q(k,n,j)$ is defined as $m$,
and if $i_{k,n}$ is not illuminated by any pixels of $P_{j}$,
$q(k,n,j)$ is defined as $0$.
In the example of Fig.~\ref{fig:variables},
$q(1,2,1)=2$ 
and $q(1,2,2)=1$
since $i_{1,2}$ is illuminated by $p_{1,2}$
and $p_{2,1}$.
$q(1,1,2)=0$ since $i_{1,1}$ is not illuminated by $P_2$.

Let us define two imaginary pixels $p_{1,0}=p_{2,0}=0$
to simplify formulas.
Then, 
using these definitions,  the constraints of the projections
are expressed as follows: 
\begin{align}
i_{k,n}= \frac{(d_{k,n,1})^2}{{\bf L}_{k,n,1}\cdot {\bf N}_{k,n}} 
p_{1,q(k,n,1)}
+
\frac{(d_{k,n,2})^2}{{\bf L}_{k,n,2}\cdot {\bf N}_{k,n}}
p_{2,q(k,n,2)}.
\label{lineareqn1}
\end{align}
where $d_{k,n,j}$ is the distance between a pixel on the object and the 
projector in order to compensate for the light fall-off and ${\bf 
L}_{k,n,j}\cdot {\bf N}_{k,n}$ is the angle between the surface normal ${\bf N}$ of $i_{k,n}$ 
and the incoming light vector ${\bf L}$ at pixel $n$ from ${\bf P}_j$ to compensate 
the Lambertian reflectance of the matte plane. 
If $q(k,n,j)=0$, then we define $d_{k,n,j}=0$ and, ${\bf L}_{k,n,j}={\bf N}_{k,n}=(1,0,0)$.

By collecting these equations, linear equations
\begin{align}
{\bf i}_1 = {\bf A}_{1,1}{\bf p}_1 + {\bf A}_{1,2}{\bf p}_2\\  
{\bf i}_2 = {\bf A}_{2,1}{\bf p}_1 + {\bf A}_{2,2}{\bf p}_2  
\label{lineareqn2}
\end{align}
follow, where ${\bf p}_j$ is a vector $[p_{j,1},p_{j,2},\cdots,p_{j,M}]$, and 
${\bf i}_k$ is a vector $[i_{k,1},i_{k,2},\cdots,i_{k,N}]$, and the matrix ${\bf 
A}_{k,j}$ is defined by its $(m,n)$-elements as 
\begin{align}
{\bf A}_{k,j}(n,m)= 
  \begin{cases}
    \frac{(d_{k,n,j})^2}{{\bf L}_{k,n,j}\cdot {\bf N}_{k,n}} & ( q(k,n,j)=m )\\
    0 & ( otherwise )
  \end{cases}
,
\label{lineareqn3}
\end{align}

\noindent By using ${\bf i} \equiv \left[ 
\begin{array}{r} 
{\bf i}_1\\
{\bf i}_2
\end{array}
\right]
$
,
$
{\bf p} \equiv 
\left[ 
\begin{array}{r} 
{\bf p}_1\\
{\bf p}_2
\end{array}
\right]
$
and
$
{\bf A} \equiv 
\left[ 
\begin{array}{rr} 
{\bf A}_{1,1}&{\bf A}_{1,2}\\
{\bf A}_{2,1}&{\bf A}_{2,2}
\end{array}
\right]
$,
we get our complete linear system
\begin{align}
{\bf i} = {\bf A}{\bf p}.
\label{lineareqn4}
\end{align}

\begin{figure}[tb]
	\centering
\includegraphics[height = 27mm]{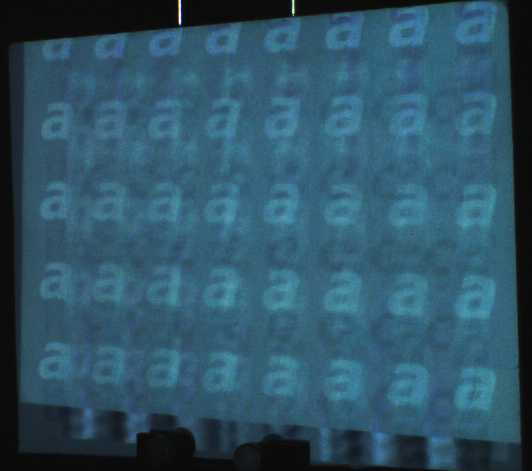}
\includegraphics[height =  27mm]{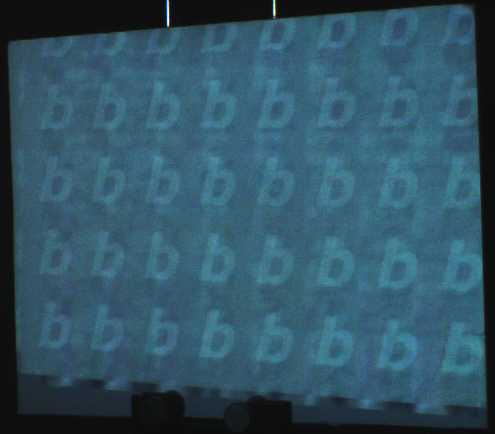}
\includegraphics[height =  27mm]{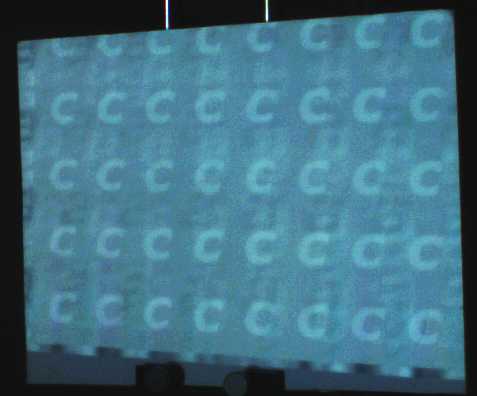}
\caption{Simultaneous pattern projection at three different depths.}
\vspace{-3mm}
\label{fig:simple_test}
\end{figure}

This system can be solved with linear algebra techniques as in~\cite{Scarzanella:psivt15}.
%\fnoteII{
Simple patterns are used to confirm that our algorithm can simultaneously project more than two planes as shown in Fig.~\ref{fig:simple_test}.
%}

In this paper, we assume there are no defocus blur of the projectors, 
which is true only if the surfaces of 3D screens are near the focal planes.
In case that the defocus blur is not neglectable, 
the projected images becomes a convolution of the original image and the blur kernel. 
Let us assume a typical setup, 
where we fix a plane in 3D space, 
place two projectors with the same aperture size at the same distances from the 
fixed plane, 
and adjust the focuses of the two projectors 
onto 
%so that the focal planes of them
%becomes the same as 
the fixed plane.
Note that these conditions are often approximately fulfilled in real setup. 
In this setup, the sizes of the blur kernel becomes the same for both projectors,
even for the off-focus surfaces,
thus, the projected images from both the projectors are convolved by the same blur kernel. 
Then, because of distributivity property of convolutions, 
the added image becomes the convolution of the non-blurred added image and the blur kernel.
This means that, in this case, the off-focus blur on each projected images does not 
%\fnoteII{
disturb the image addition process of equation (\ref{lineareqn4}), 
but only blur the resulting projected images. 
%}
In reality, 
we did not find a high divergence between the simulations with non-blur assumptions 
and the real experiments with blur.

\subsection{Problem reduction using epipolar constraints and constraint-aware optimization}

The problem to be solved is to obtain ${\bf p}$ given ${\bf i}$ and ${\bf A}$. 
The length of vector ${\bf p}$ is $M\cdot J$, while the length of vector ${\bf 
i}$ is $N\cdot K$. 
The matrix ${\bf A}$ is a very large sparse matrix. To 
model the real system, this simple linear model has two problems. 
First, it implies a global solution through pseudo-inversion of a very large matrix. Second, since ${\bf i}$ 
and ${\bf p}$ are images, their elements should be non-negative values with a 
fixed dynamic range. However, the lack of positivity constraints in the solution 
of the sparse system means that ${\bf p}$ may include negative or very large positive elements. 
This was solved in \cite{Scarzanella:psivt15}~ by normalising ${\bf p}$ so that the elements are in the range of [0,255]
using a sparse matrix linear algebra solver.
However, the effect of this is a compression of the resulting 
dynamic range and a lowering of the contrast.

\begin{figure}[t]
	\centering
\includegraphics[width = .42\textwidth]{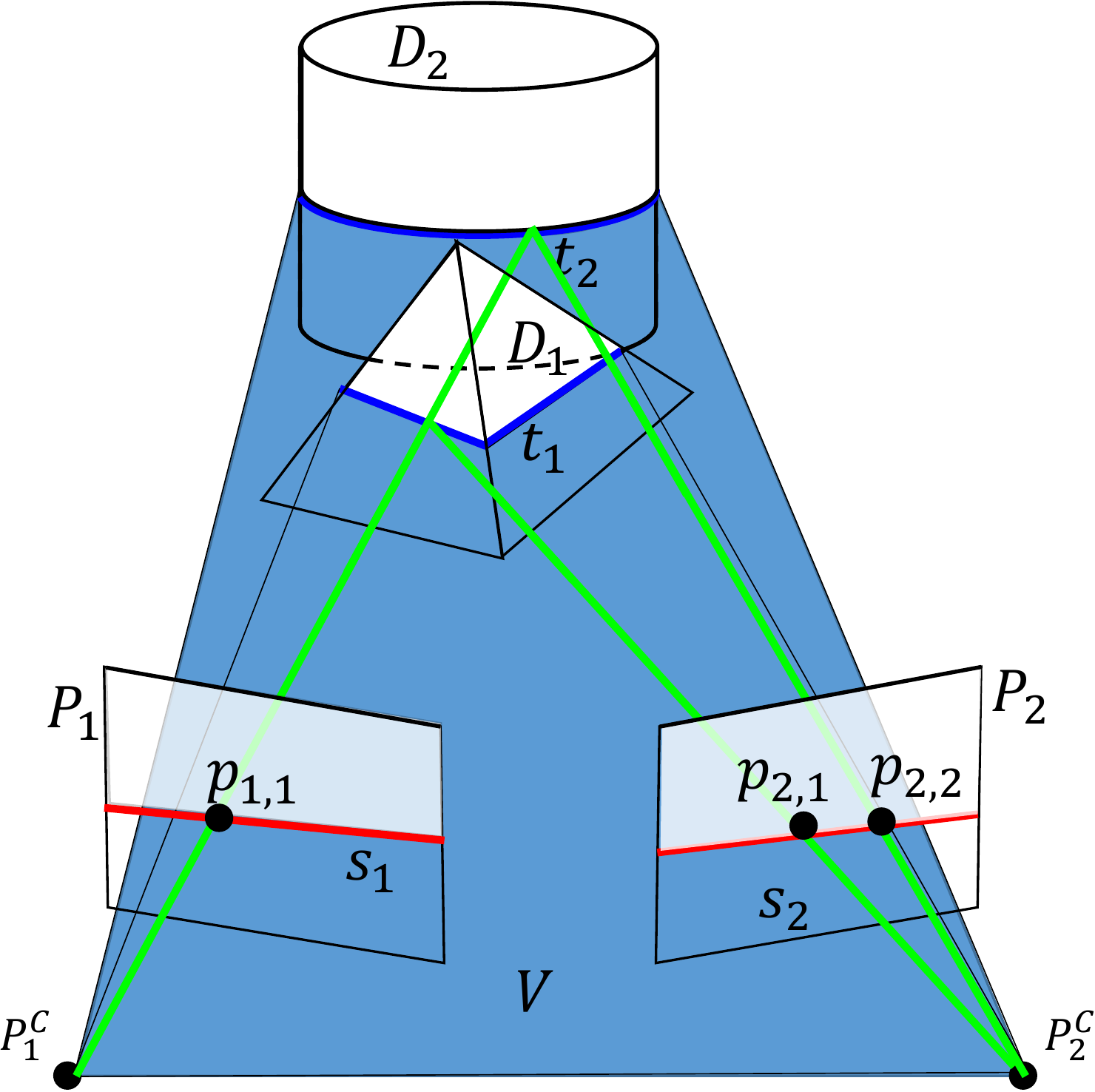}
\caption{Epipolar relationship between projectors for the proposed optimization.}
\label{Fig_epi}
%\end{figure}
%%\begin{figure}[t]
%	\centering
%%	\includegraphics[width=0.7\textwidth]{graycode_depth.eps}
%	\includegraphics[width=0.7\textwidth]{graycode_depth-eps-converted-to.pdf}
%	\caption{Correspondences between objects.}
%	\label{fig:correspond}
%%	\vspace{-2mm}
%	\vspace{-2mm}
\end{figure}

Here, we consider the case that $J=2$, thus, we assume the two projector scenario shown in Fig. \ref{Fig_epi} with two objects $D_1$ and $D_2$ onto which the images are to be projected. 
%In this case, we can greatly reduce the size of the problem. 
%In order to better explain our proposed optimization scheme, 
%Let us consider the geometric setup shown in Fig. \ref{Fig_epi} with two objects $D_1$ and $D_2$ as 3D screens.
Given the optical centers of the two projectors $P_1^C$ and $P_2^C$ as well as any pixel on the first pattern $p_{1,1}$ (without loss of generality), 
the epipolar plane $V$ defined by the three points will intersect projected patterns ${\bf P}_1$ and ${\bf P}_2$ at lines $s_1$, $s_2$,
and 3D screens $D_1$ and $D_2$ at $t_1$ and $t_2$, respectively.
We can see that the pixel compensation between ${\bf P}_1$ and ${\bf P}_2$
occurs only within epipolar lines $s_1$ and $s_2$.
This means that the problems of optimizing the pixels of the projected images can be solved for each epipolar line,
instead of solving entire projected pixels of ${\bf P}_1$ and ${\bf P}_2$.

% between projectors ${\bf P}_1$ and ${\bf P}_2$.

To obtain a finite set of the optimized pixels, we use the following steps. 
For any pixel $p_{1,1}$  along $s_1$, this will correspond to points along the intersections $t_1$, $t_2$ of the epipolar plane with the image planes. 
Similarly, 
 %projecting any point from $t_1$, $t_2$ will map to a point along $s_2$. 
the projected pixels of these points to ${\bf P}_2$ along $s_2$,
$p_{2,1}$ and $p_{2,2}$  in Fig. \ref{Fig_epi}, 
are added  to the list of variables.
%Therefore, by 
By iterating the process, 
%we can see that the sequence of compensating patterns involving the point $p$ consists of a finite number of steps involving points 
we obtain the list of variables involved in the calculation of the pixel compensation with respect to $s_1$ and $s_2$.
%along the intersections of the epipolar plane with the projector and the scene geometry. 
Moreover, the spacing between points in the sequence on each intersection line will depend on the distance and the angle of vergence between the two projectors.

\begin{figure}[t]
\centering
%\small
\scriptsize
\setlength{\tabcolsep}{5pt}
\begin{subfigure}{0.59\textwidth}
    \begin{tabular}{rrrr}    \toprule
    \multicolumn{4}{c}{PSNR} \\
    \midrule
		\centering
    & LF & $EO_{-100}^{255}$ & $EO_0^{255}$ \\
    \textit{Lena/Mandrill} & 10.9 / 9.2 & 10.8 / 11.3 & \textbf{19.6 / 20.9} \\
    \textit{Lena/Peppers} & 10.4 / 7.8 & 11.2 / 10.7 & \textbf{19.4 / 20.4} \\
    \textit{Peppers/House} & 8.0 / 8.2 & 9.4 / 11.1 & \textbf{18.3 / 20.0} \\
    \textit{Peppers/Lena} & 8.7 / 8.0 & 10.3 / 10.8 & \textbf{19.7 / 20.0} \\
%    \textit{Lena/Mandrill} & 10.867 / 9.218 & 10.783 / 11.275 & \textbf{19.588 / 20.863} \\
%    \textit{Lena/Peppers} & 10.409 / 7.750 & 11.226 / 10.653 & \textbf{19.414 / 20.385} \\
%    \textit{Peppers/House} & 7.965 / 8.188 & 9.394 / 11.113 & \textbf{18.257 / 19.996} \\
%    \textit{Peppers/Lena} & 8.723 / 7.981 & 10.349 / 10.829 & \textbf{19.663 / 20.027} \\
    \bottomrule
    \end{tabular}%
\end{subfigure}
\begin{subfigure}{0.39\textwidth}
\begin{tabular}{rrrr}
    \toprule
    \multicolumn{3}{c}{SSIM} \\
    \midrule
		\centering
    LF & $EO_{-100}^{255}$ & $EO_0^{255}$ \\
    -0.04 / 0.20 & 0.04 / 0.37 & \textbf{0.66 / 0.71} \\
    -0.03 / -0.09 & 0.07 / 0.24 & \textbf{0.71 / 0.73} \\
    -0.02 / 0.11 & 0.11 / 0.14 & \textbf{0.58 / 0.67} \\
    0.02 / -0.16 & 0.20 / 0.02 & \textbf{0.77 / 0.68} \\
%    -0.038 / 0.196 & 0.041 / 0.365 & \textbf{0.659 / 0.709} \\
%    -0.025 / -0.089 & 0.068 / 0.238 & \textbf{0.705 / 0.733} \\
%    -0.020 / 0.109 & 0.113 / 0.138 & \textbf{0.580 / 0.669} \\
%    0.024 / -0.163 & 0.202 / 0.0218 & \textbf{0.769 / 0.679} \\
    \bottomrule
    \end{tabular}
\end{subfigure}
\caption{Top: numerical PSNR and SSIM results for validation of the system prototype according to the method tested. 
For each data entry, 
the accuracy measures are indicated for each depth plane individually in the form P/Q, 
where P and Q are the accuracy measures at the first and second position respectively. Best performance is highlighted in bold.}
\label{Fig_realTableHist}
\end{figure}

Hence, instead of formulating the problem as a large global optimization, we decompose it into a series of small local problems of the form:
\begin{equation}
\check{\textbf{i}} =  \check{\textbf{A}} \check{\textbf{p}},
\end{equation}
\noindent where $\check{\textbf{i}}$ and $\check{\textbf{p}}$ are the image and pattern pixels related by epipolar geometry and mapped to each other by the ray-tracing matrix $\check{\textbf{A}}$. Contrarily to the large sparse system, the number of elements involved in each local optimisation is $10^1 - 10^2$ depending on the projector setup. This allows us to expand the above equation into a series of explicit sums:

%\begin{equation}
%\begin{cases}
%\check{\textbf{i}}_1 = \sum\limits_{k=1}^N \check{\textbf{A}}_{1,k} \check{\textbf{p}}_k\\
%\check{\textbf{i}}_2 = \sum\limits_{k=1}^N \check{\textbf{A}}_{2,k} \check{\textbf{p}}_k\\
%\vdots\\
%\check{\textbf{i}}_M = \sum\limits_{k=1}^N \check{\textbf{A}}_{M,k} \check{\textbf{p}}_k
%\end{cases},
%\end{equation}
\begin{equation}
\check{\textbf{i}}_j = \sum\limits_{k=1}^N \check{\textbf{A}}_{j,k} \check{\textbf{p}}_k, 
\end{equation}
\noindent where 
%$1 \le j \le M$, 
$1 \le j \le M$,
$M$ is the number of points on the image planes,
 and $N$ is the number of points from the patterns involved in this sequence. For each one of these explicit sums, we solve for the optimal pattern pixels $\check{\textbf{p}}_i^*$ by solving the constrained optimisation problem:

\begin{equation}
 \check{\textbf{p}}_i^* = \min_{\check{\textbf{p}}_i} \sum\limits_{k=1}^N \left(\check{\textbf{i}}_j - \check{\textbf{A}}_{j,k} \check{\textbf{p}}_k\right)^2 \mid \check{\textbf{p}}_k \in [a,b],
\end{equation}
\noindent where $[a,b]$ is the allowed range of intensities during the pattern generation. With this strategy, we are able to independently solve the pattern generation problem for small sets of points at a time, which in turn allows us to impose constraints and to confidently optimise each chain with fewer chances of getting stuck in a local minimum. 

\begin{figure}[t]
\centering
\begin{tabular}{ccccc}
\begin{subfigure}{0.22\textwidth}
    \includegraphics[trim={5cm 3cm 5cm 3cm},clip=true,width = 1\textwidth]{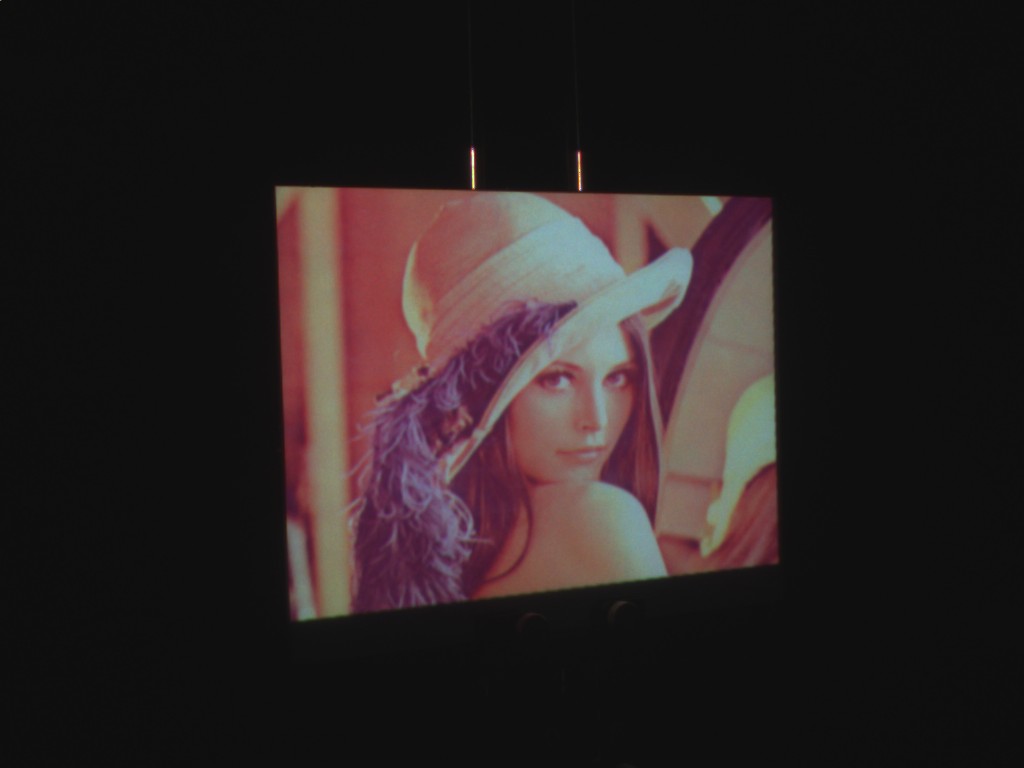}
%   \FramedBox{3.5cm}{1\textwidth}{Original image 1.}
\end{subfigure}
&
\begin{subfigure}{0.22\textwidth}
    \includegraphics[trim={2cm 5cm 13cm 5cm},clip = true,width = 1\textwidth]{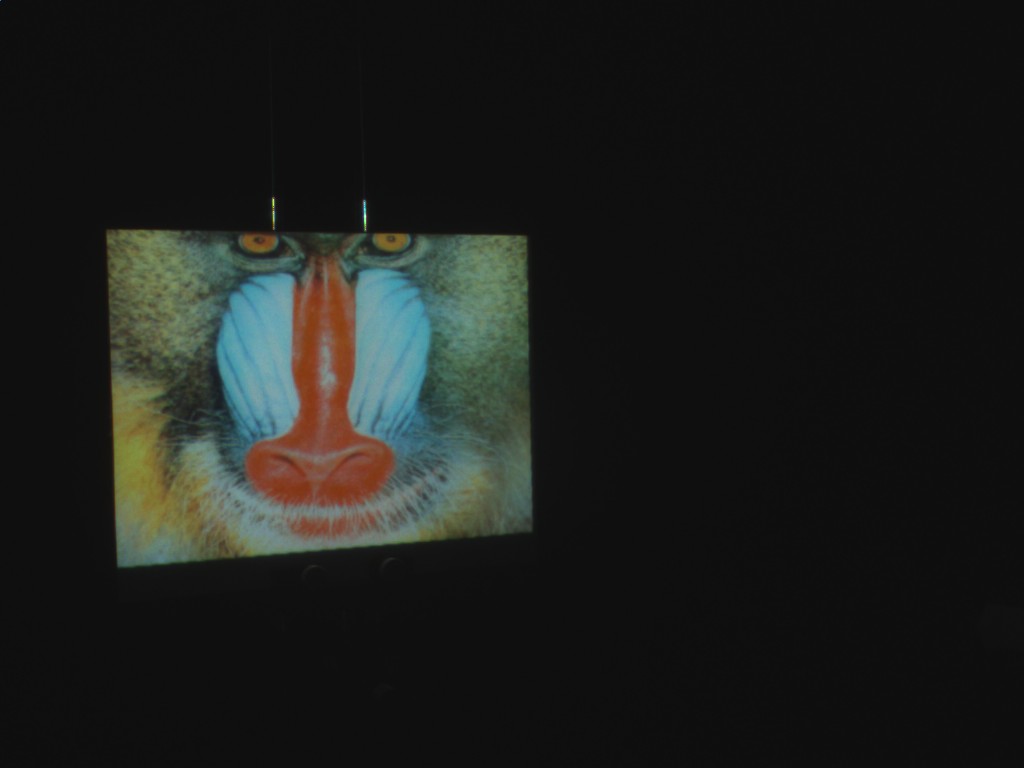}
%    \FramedBox{3.5cm}{1\textwidth}{Original image 2.}
\end{subfigure}
&
\begin{sideways}
 \hspace{-.8cm}
 \Large Original
\end{sideways}
&
\begin{subfigure}{0.22\textwidth}
    \includegraphics[trim={5cm 3cm 5cm 3cm},clip=true,width = 1\textwidth]{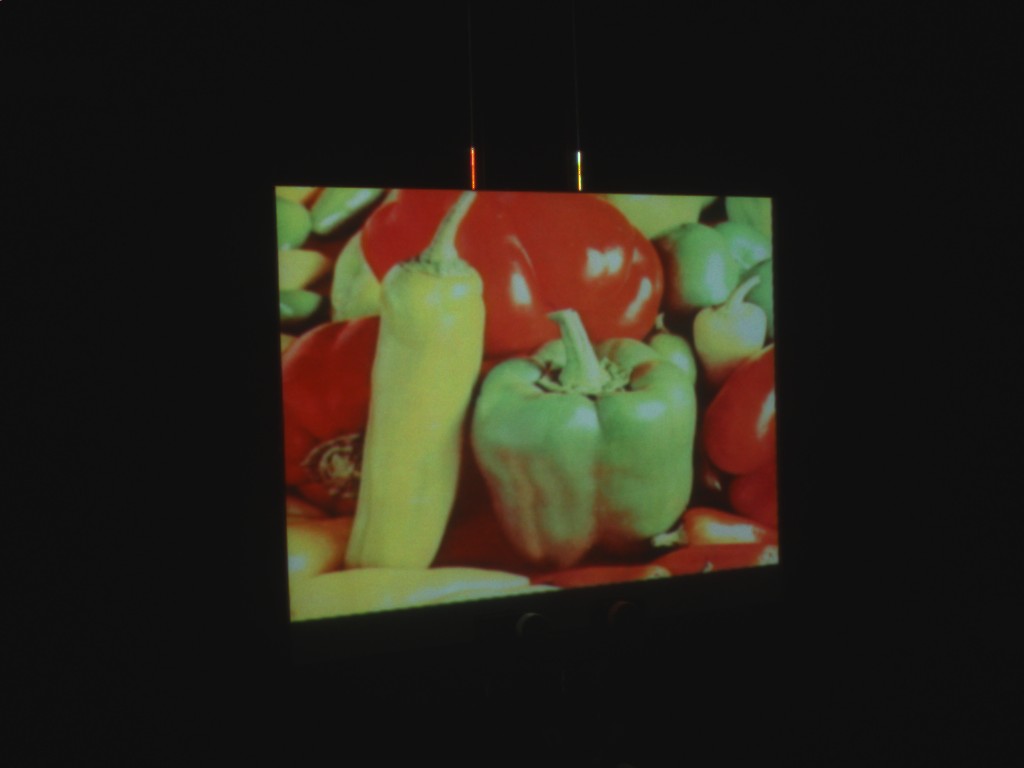}
%    \FramedBox{3.5cm}{1\textwidth}{Original image 3.}
\end{subfigure}
&
\begin{subfigure}{0.22\textwidth}
    \includegraphics[trim={2cm 5cm 13cm 5cm},clip = true,width = 1\textwidth]{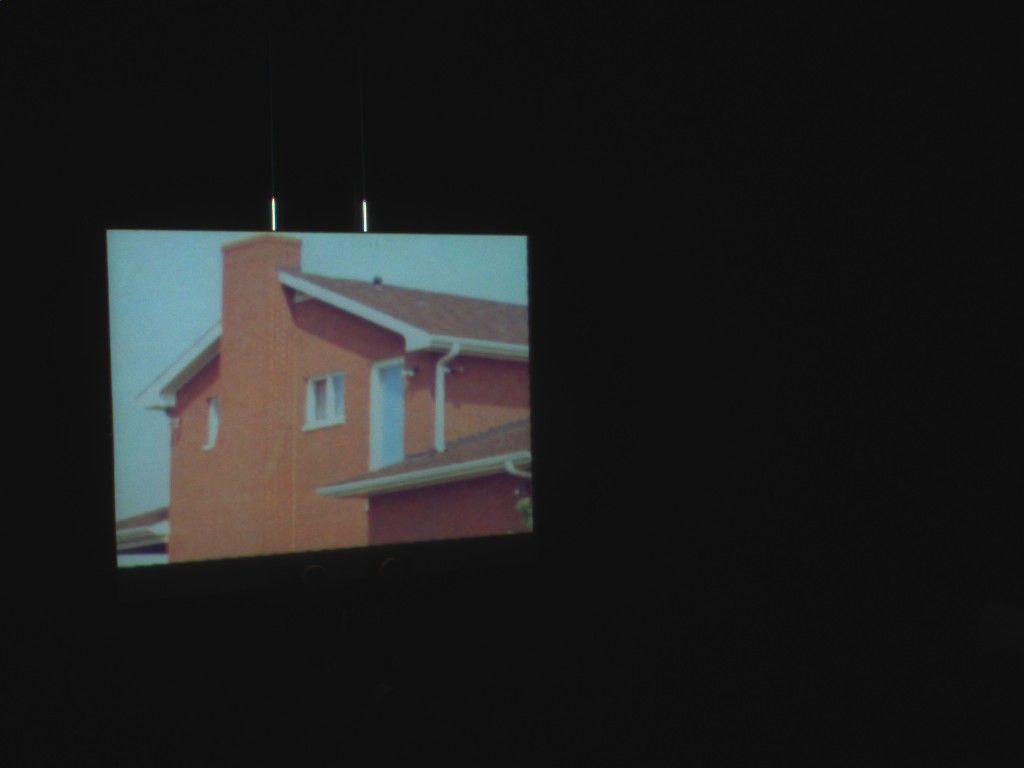}
%    \FramedBox{3.5cm}{1\textwidth}{Original image 4.}
\end{subfigure}
\\
\begin{subfigure}{0.22\textwidth}
    \includegraphics[trim={5cm 3cm 5cm 3cm},clip=true,width = 1\textwidth]{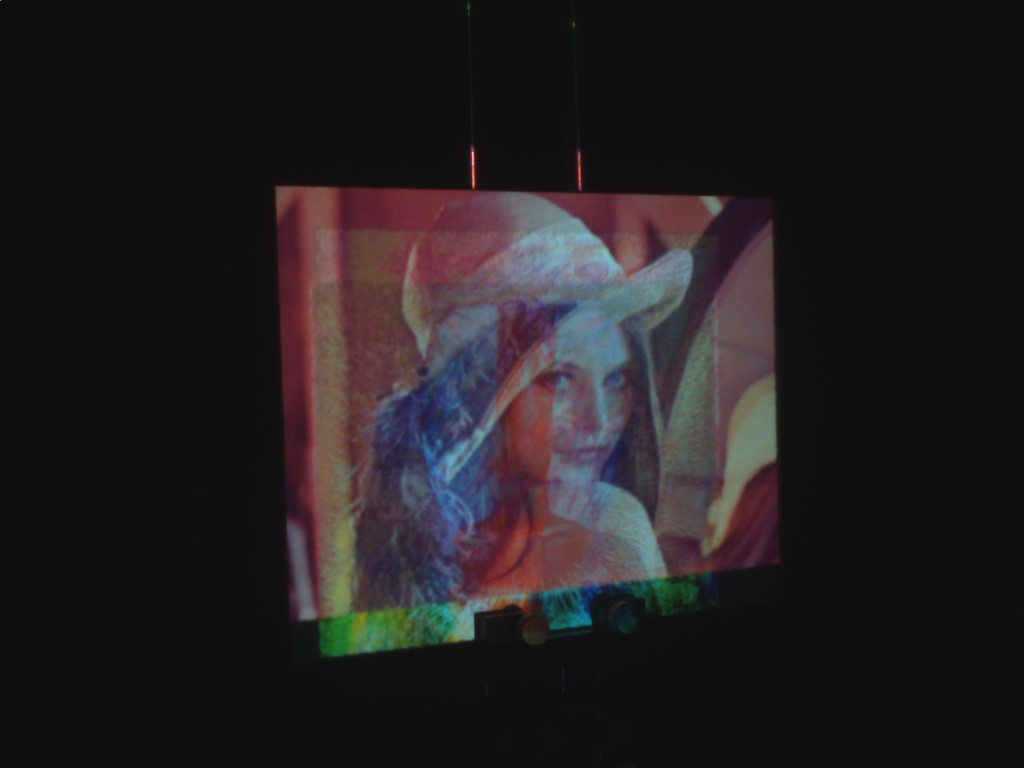}
%    \FramedBox{3.5cm}{1\textwidth}{Plane result with dense depth1.}
\end{subfigure}
&
\begin{subfigure}{0.22\textwidth}
    \includegraphics[trim={2cm 5cm 13cm 5cm},clip = true,width = 1\textwidth]{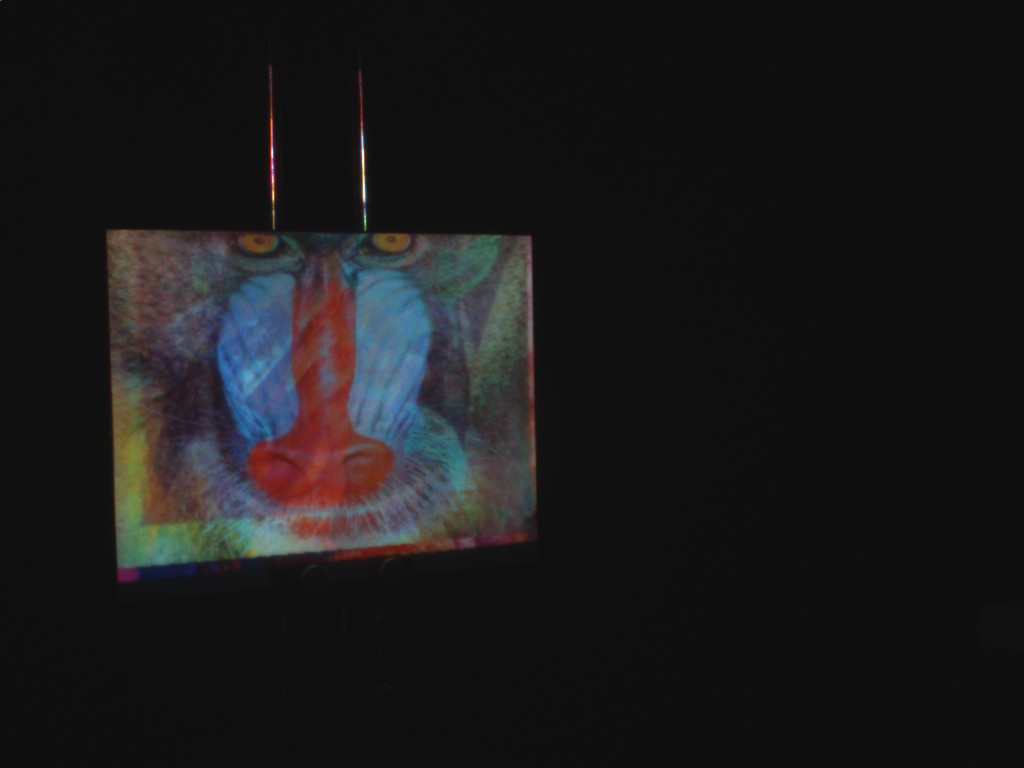}
 %   \FramedBox{3.5cm}{1\textwidth}{Plane result with dense depth2.}
\end{subfigure}
&
\begin{sideways}
 \hspace{-.8cm}
 \Large $EO_0^{255}$% $\rightarrow$ 
\end{sideways}
&
\begin{subfigure}{0.22\textwidth}
    \includegraphics[trim={5cm 3cm 5cm 3cm},clip=true,width = 1\textwidth]{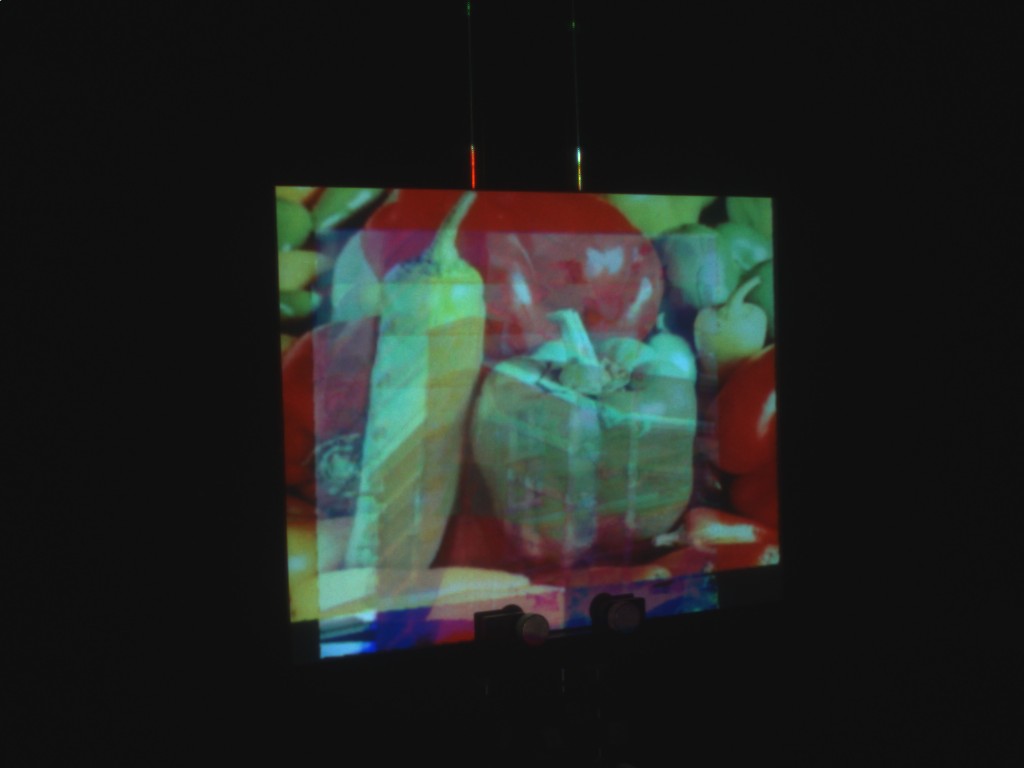}
%    \FramedBox{3.5cm}{1\textwidth}{Plane result with dense depth1.}
\end{subfigure}
&
\begin{subfigure}{0.22\textwidth}
    \includegraphics[trim={2cm 5cm 13cm 5cm},clip = true,width = 1\textwidth]{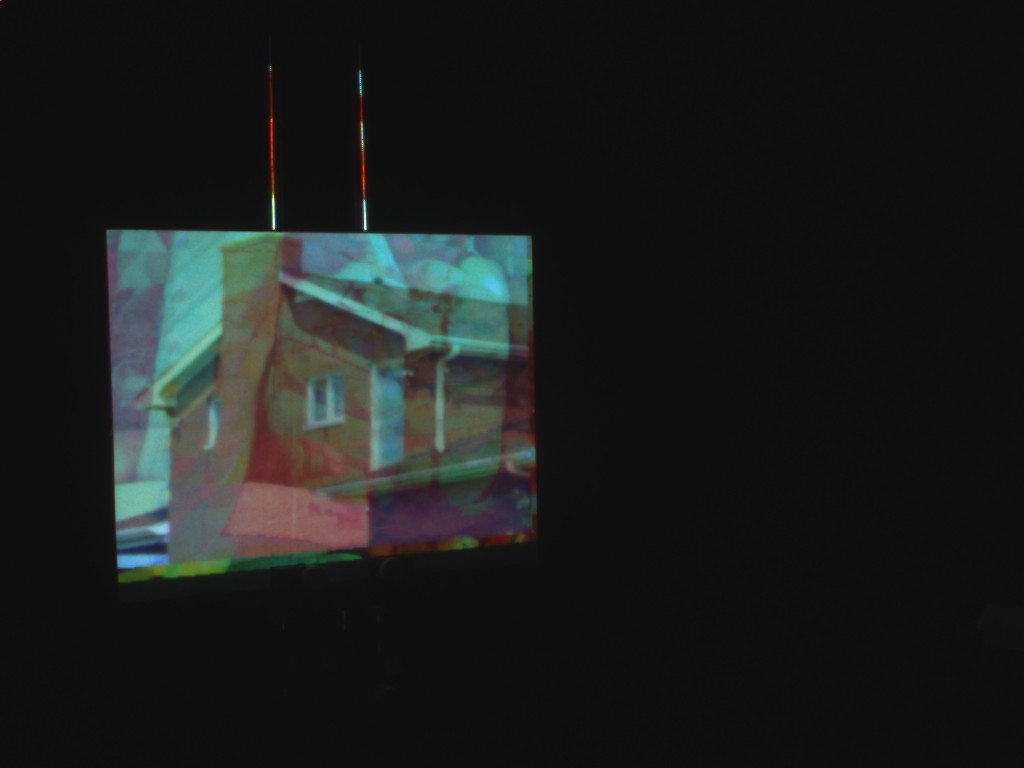}
%    \FramedBox{3.5cm}{1\textwidth}{Plane result with dense depth2.}
\end{subfigure}
\\
\begin{subfigure}{0.22\textwidth}
    \includegraphics[trim={5cm 3cm 5cm 3cm},clip=true,width = 1\textwidth]{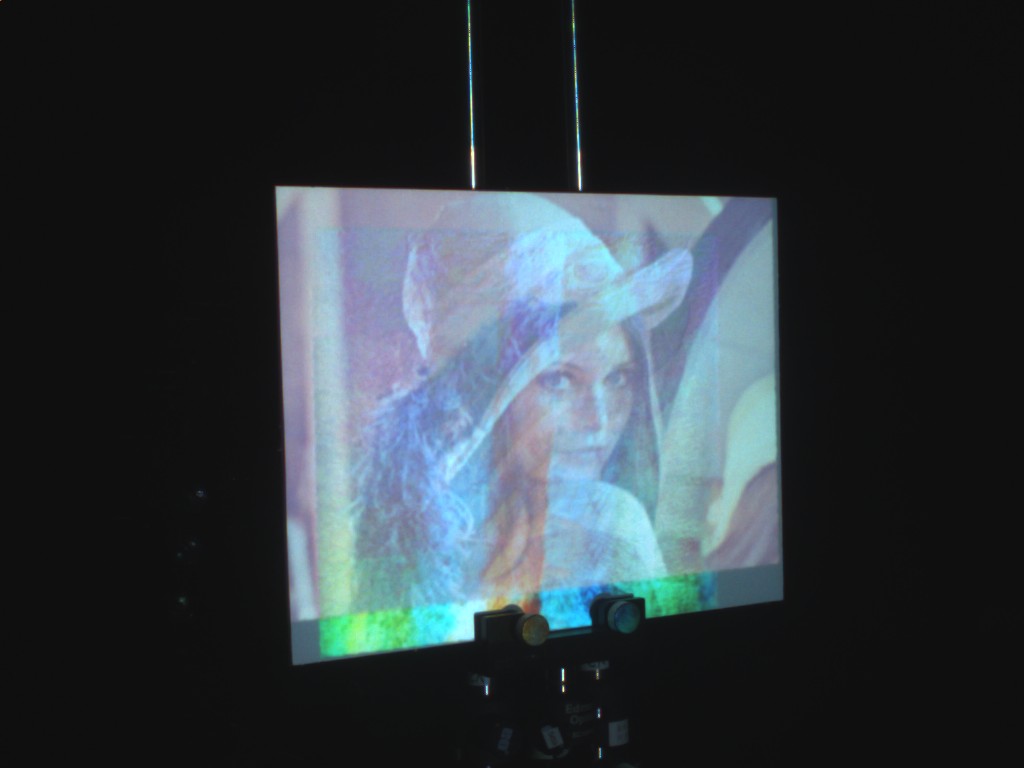}
%    \FramedBox{3.5cm}{1\textwidth}{Plane result with sparse depth1.}
\end{subfigure}
&
\begin{subfigure}{0.22\textwidth}
    \includegraphics[trim={2cm 5cm 13cm 5cm},clip = true,width = 1\textwidth]{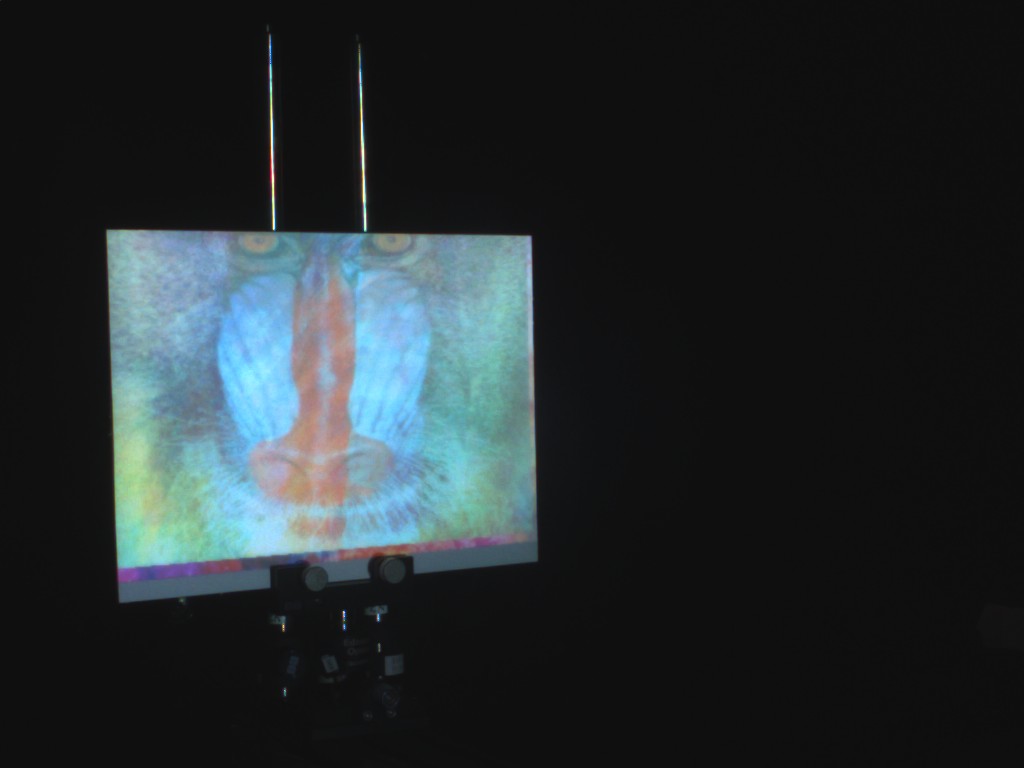}
%    \FramedBox{3.5cm}{1\textwidth}{Plane result with sparse depth2.}
\end{subfigure}
&
\begin{sideways}
 \hspace{-.8cm}
 \Large $EO_{-100}^{255}$% $\rightarrow$ 
\end{sideways}
&
\begin{subfigure}{0.22\textwidth}
    \includegraphics[trim={5cm 3cm 5cm 3cm},clip=true,width = 1\textwidth]{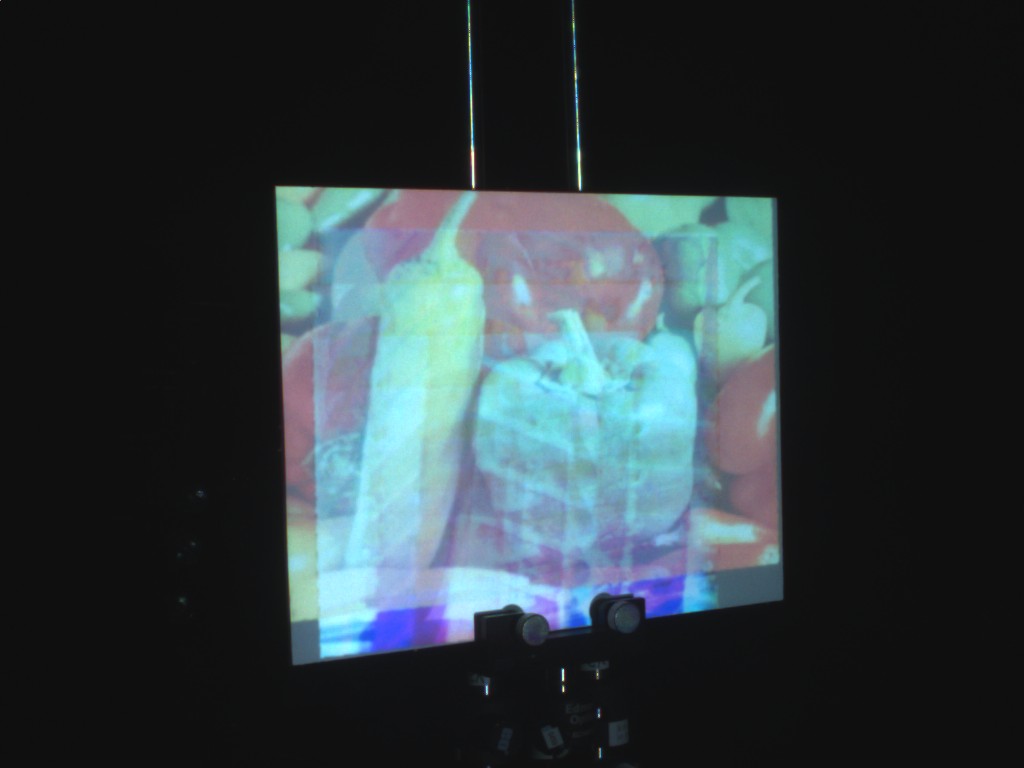}
%    \FramedBox{3.5cm}{1\textwidth}{Plane result with sparse depth1.}
\end{subfigure}
&
\begin{subfigure}{0.22\textwidth}
    \includegraphics[trim={2cm 5cm 13cm 5cm},clip = true,width = 1\textwidth]{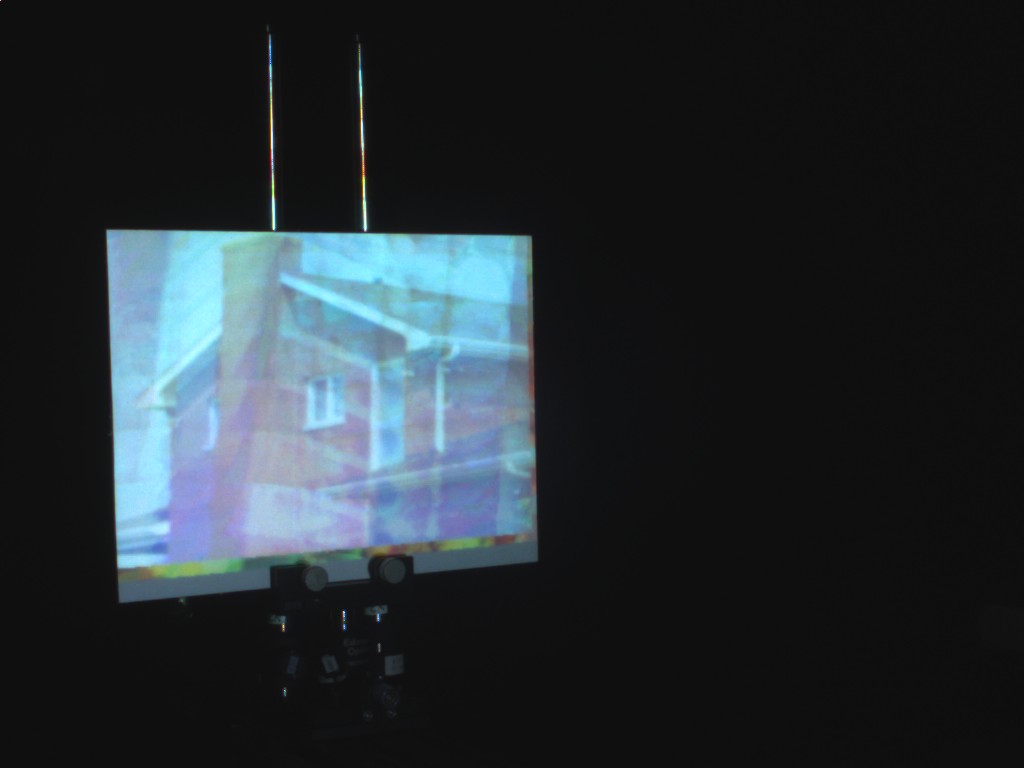}
%    \FramedBox{3.5cm}{1\textwidth}{Plane result with sparse depth2.}
\end{subfigure}
\\
\begin{subfigure}{0.22\textwidth}
    \includegraphics[trim={5cm 3cm 5cm 3cm},clip=true,width = 1\textwidth]{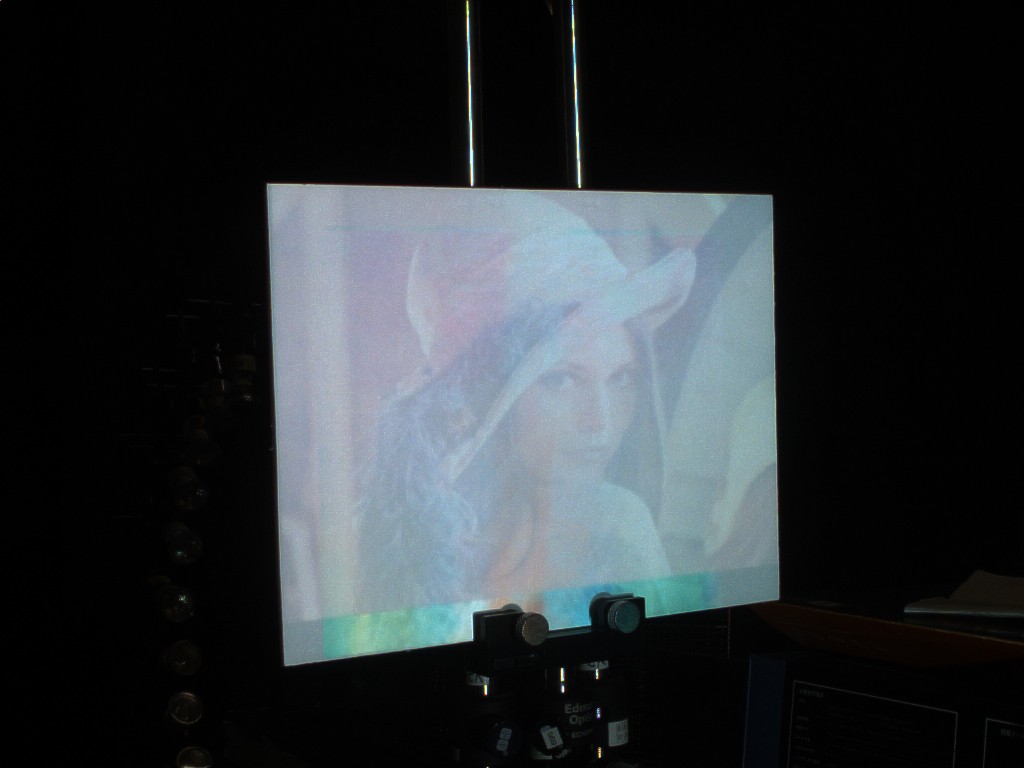}
%    \FramedBox{3.5cm}{1\textwidth}{Plane result with sparse depth1.}
\caption{Depth 1}
\end{subfigure}
&
\begin{subfigure}{0.22\textwidth}
    \includegraphics[trim={2cm 5cm 13cm 5cm},clip = true,width = 1\textwidth]{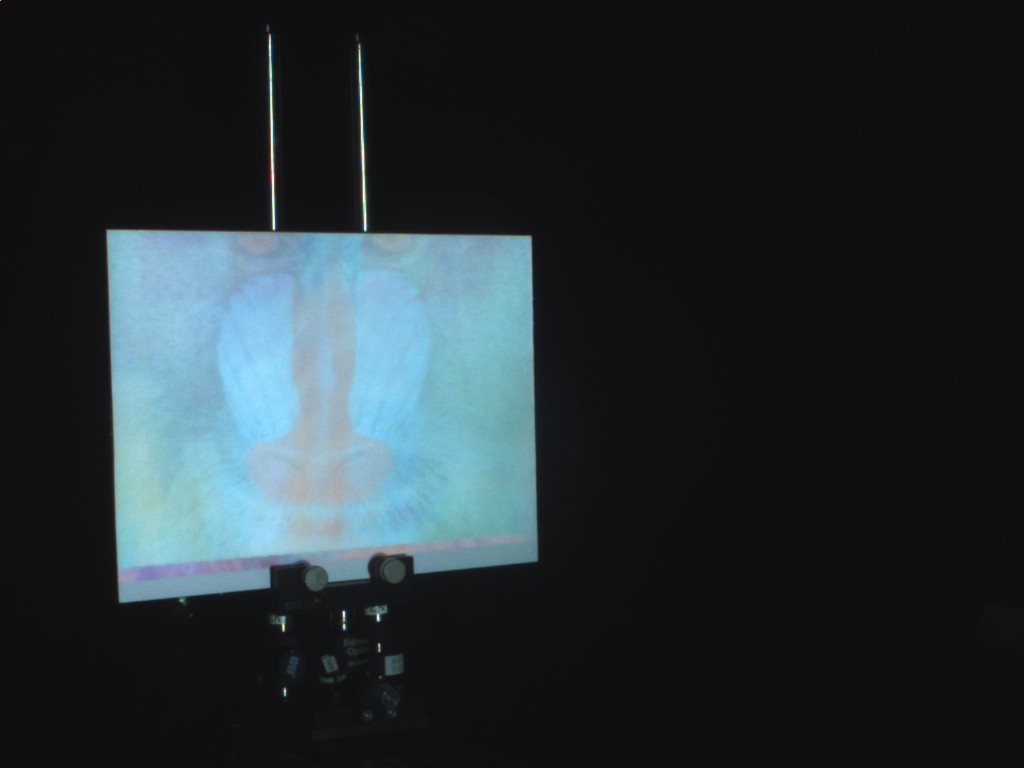}
%    \FramedBox{3.5cm}{1\textwidth}{Plane result with sparse depth2.}
\caption{Depth 2}
\end{subfigure}
&
\begin{sideways}
 %\hspace{-.8cm}
 \Large  LF
\end{sideways}
&
\begin{subfigure}{0.22\textwidth}
    \includegraphics[trim={5cm 3cm 5cm 3cm},clip=true,width = 1\textwidth]{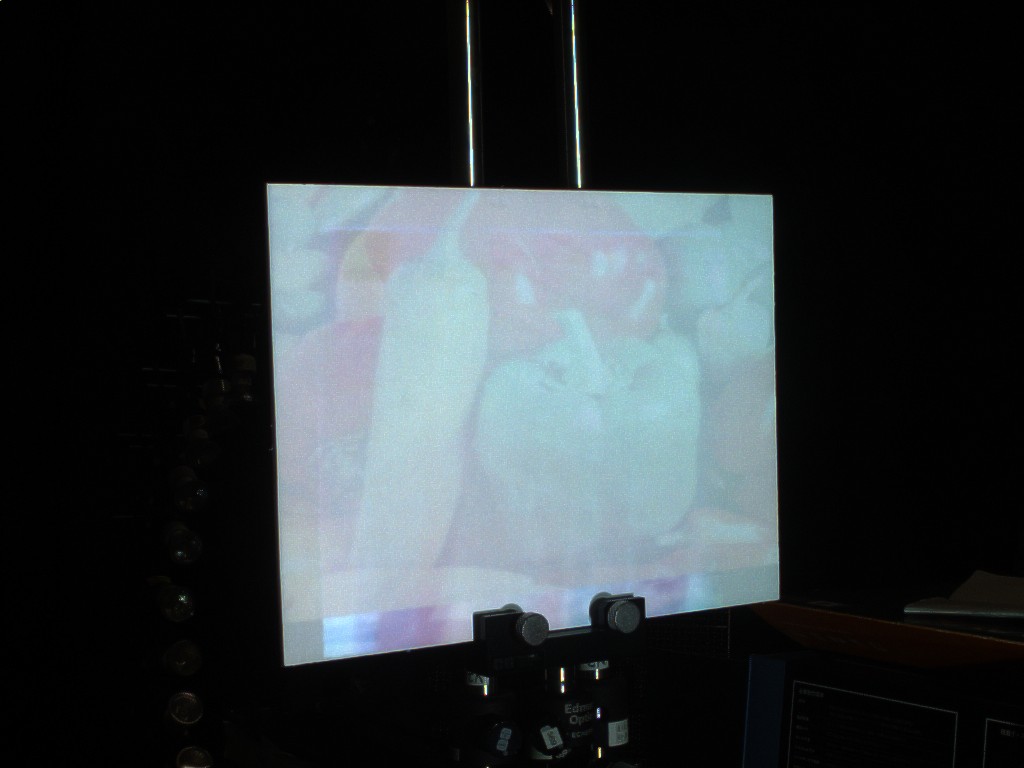}
%    \FramedBox{3.5cm}{1\textwidth}{Plane result with sparse depth1.}
\caption{Depth 1}
\end{subfigure}
&
\begin{subfigure}{0.22\textwidth}
    \includegraphics[trim={2cm 5cm 13cm 5cm},clip = true,width = 1\textwidth]{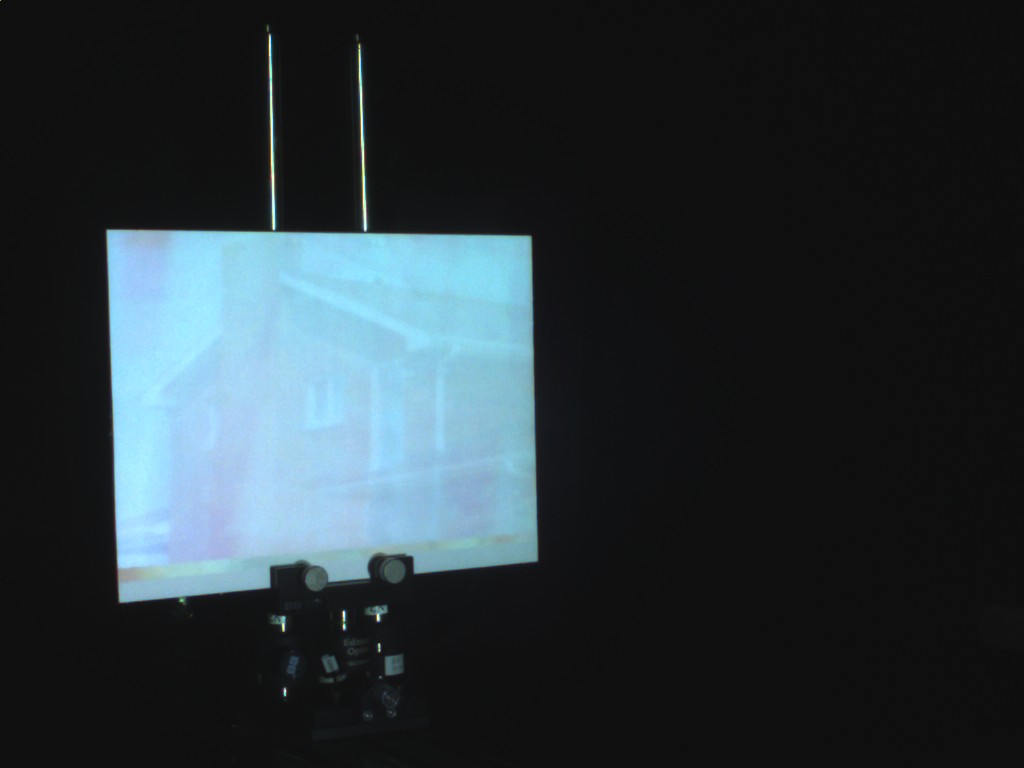}
 %   \FramedBox{3.5cm}{1\textwidth}{Plane result with sparse depth2.}
\caption{Depth 2}
\end{subfigure}
\\
\end{tabular}
%\knote{[Hirukawa] input images for comparison of sparse and dense version.}
\caption{Real projection results generated with proposed and linear factorisation algorithm. From top to bottom, left to right, the pairs \textit{Lena/Mandrill} and \textit{Peppers/House} for the methods $EO_{0}^{255}$, $EO_{-100}^{255}$ and $LF$~\cite{Scarzanella:psivt15}.} 
\label{Fig_synthResults}
\end{figure}

\begin{figure}[t]
\centering
\begin{subfigure}{0.24\textwidth}
    \includegraphics[trim={3cm 3cm 7cm 5cm}, clip = true,width = 1\textwidth]{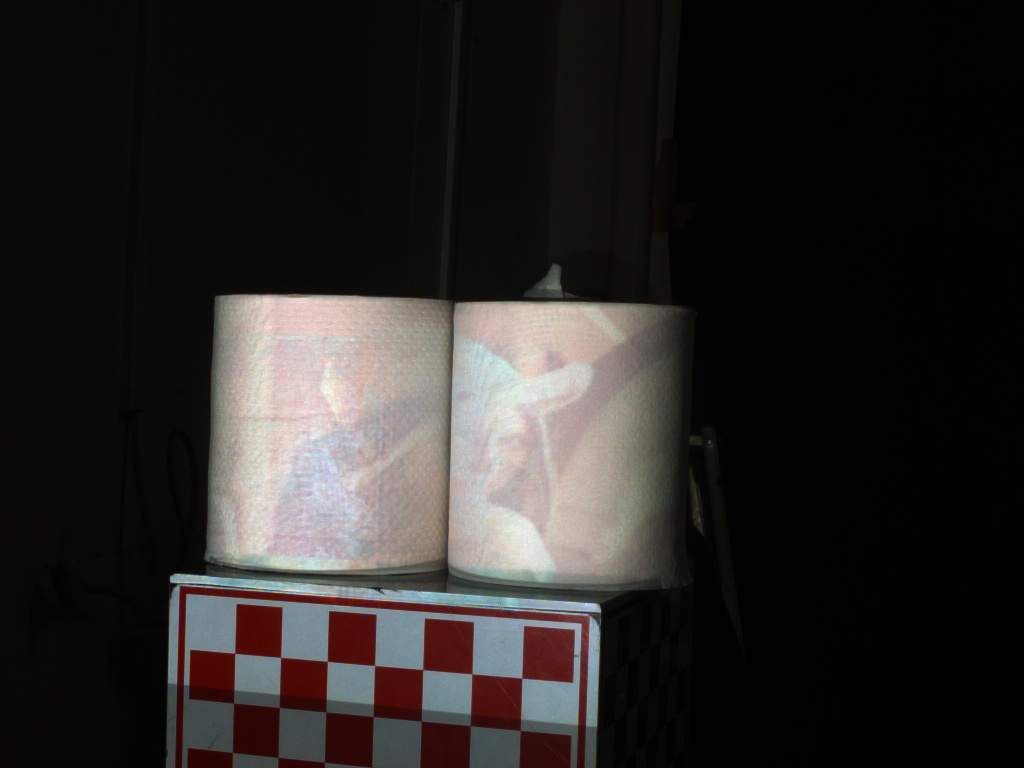}
%    \FramedBox{3.5cm}{1\textwidth}{Plane result with sparse depth1.}
\caption{}
\end{subfigure}
\begin{subfigure}{0.24\textwidth}
    \includegraphics[trim={8.9cm 5cm 8cm 8cm}, clip = true,width = 1\textwidth]{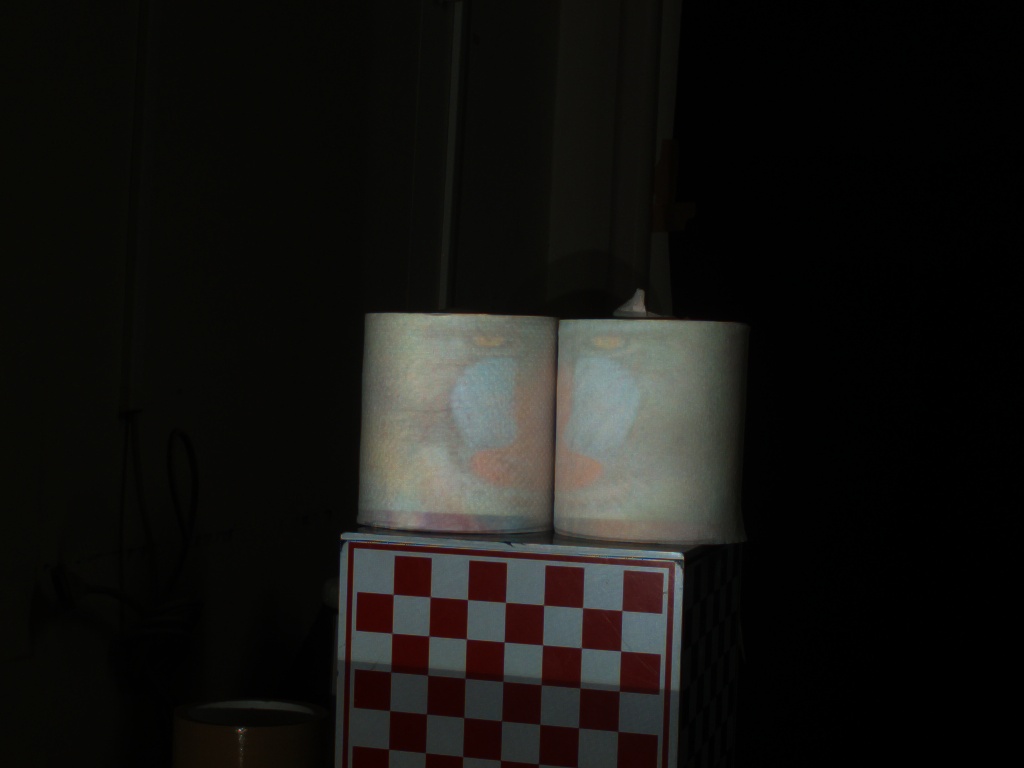}
%    \FramedBox{3.5cm}{1\textwidth}{Plane result with dense depth1.}
\caption{}
\end{subfigure}
\begin{subfigure}{0.24\textwidth}
    \includegraphics[trim={3cm 3cm 7cm 5cm}, clip = true,width = 1\textwidth]{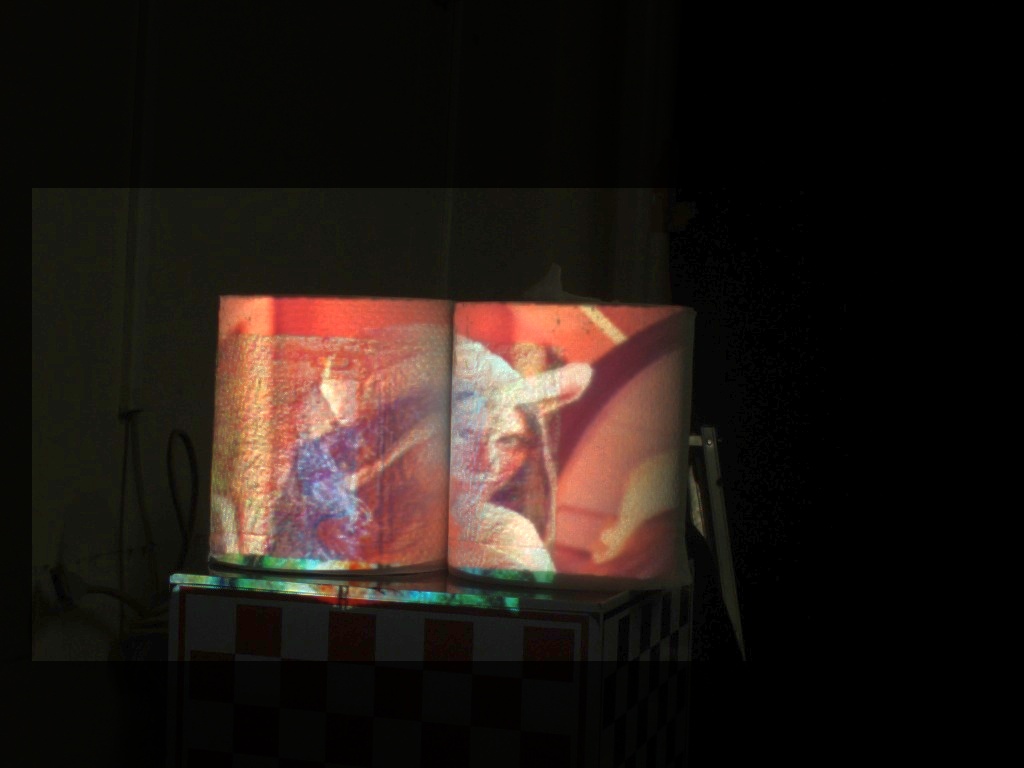}
%    \FramedBox{3.5cm}{1\textwidth}{Plane result with sparse depth2.}
\caption{}
\end{subfigure}
\begin{subfigure}{0.24\textwidth}
    \includegraphics[trim={8.9cm 5cm 8cm 8cm}, clip = true,width = 1\textwidth]{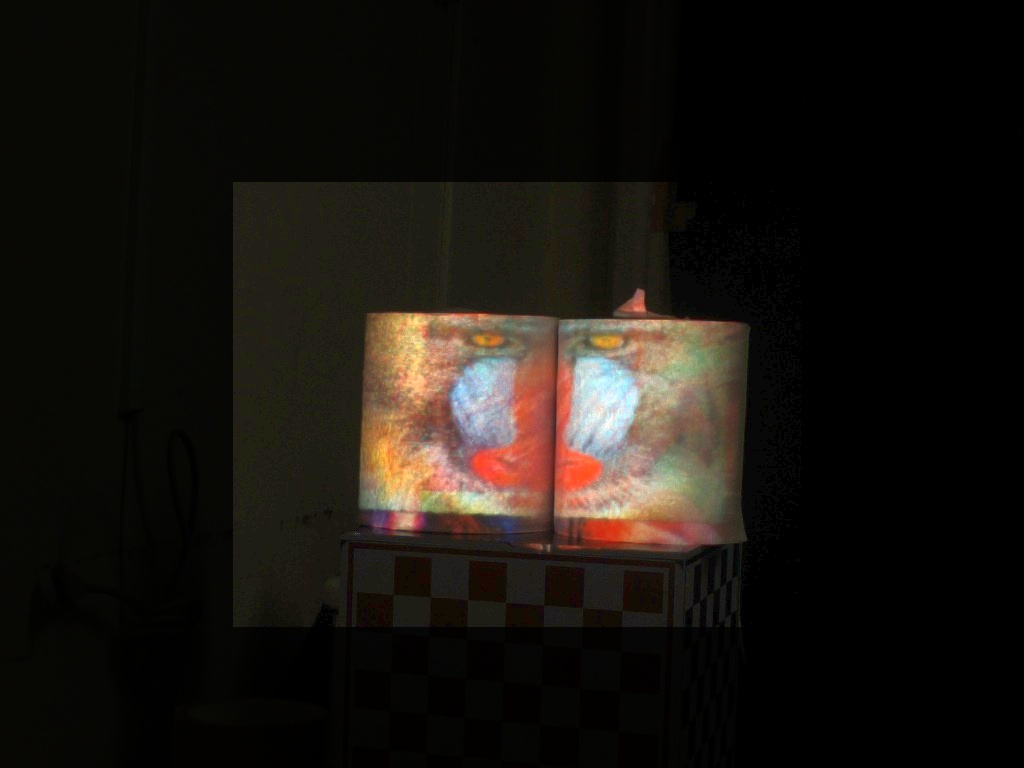}
 %   \FramedBox{3.5cm}{1\textwidth}{Plane result with dense depth2.}
\caption{}
\end{subfigure}
%\knote{[Hirukawa] input images for comparison of sparse and dense version.}
%\label{Fig_synthResults3D}
\centering
\begin{subfigure}{0.24\textwidth}
    \includegraphics[trim={0cm 2.7cm 3cm 0cm},clip = true,width = 1\textwidth]{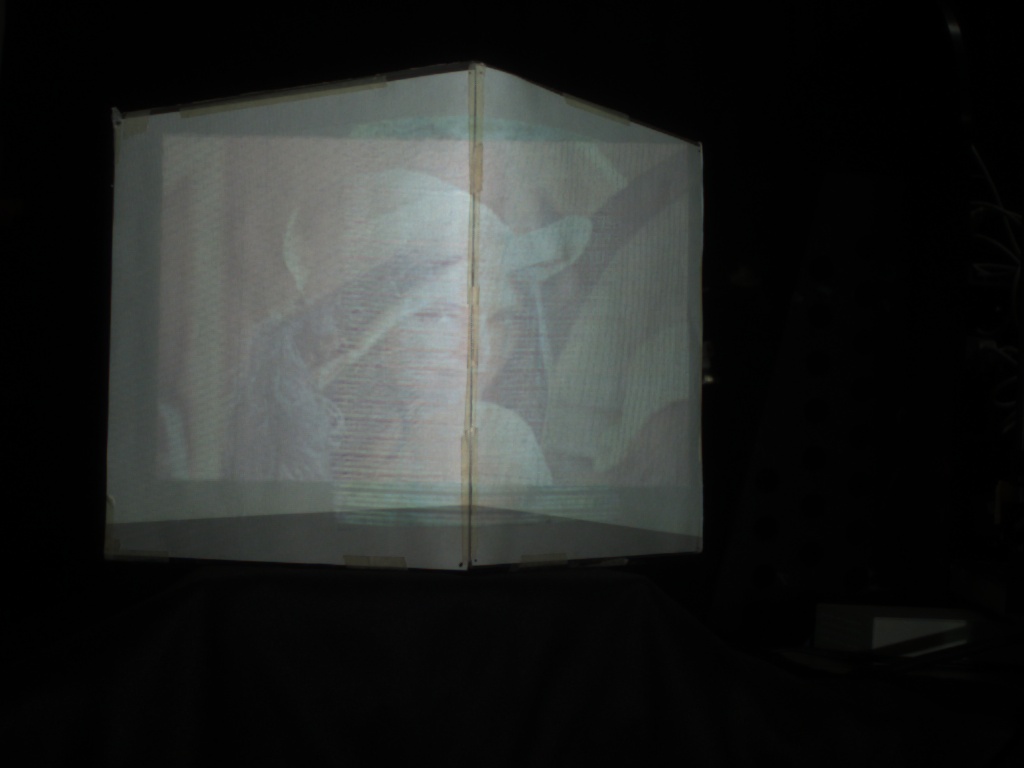}
%    \FramedBox{3.5cm}{1\textwidth}{Plane result with sparse depth1.}
\caption{}
\end{subfigure}
\begin{subfigure}{0.24\textwidth}
    \includegraphics[trim={4cm 4cm 6cm 4cm},clip = true,width = 1\textwidth]{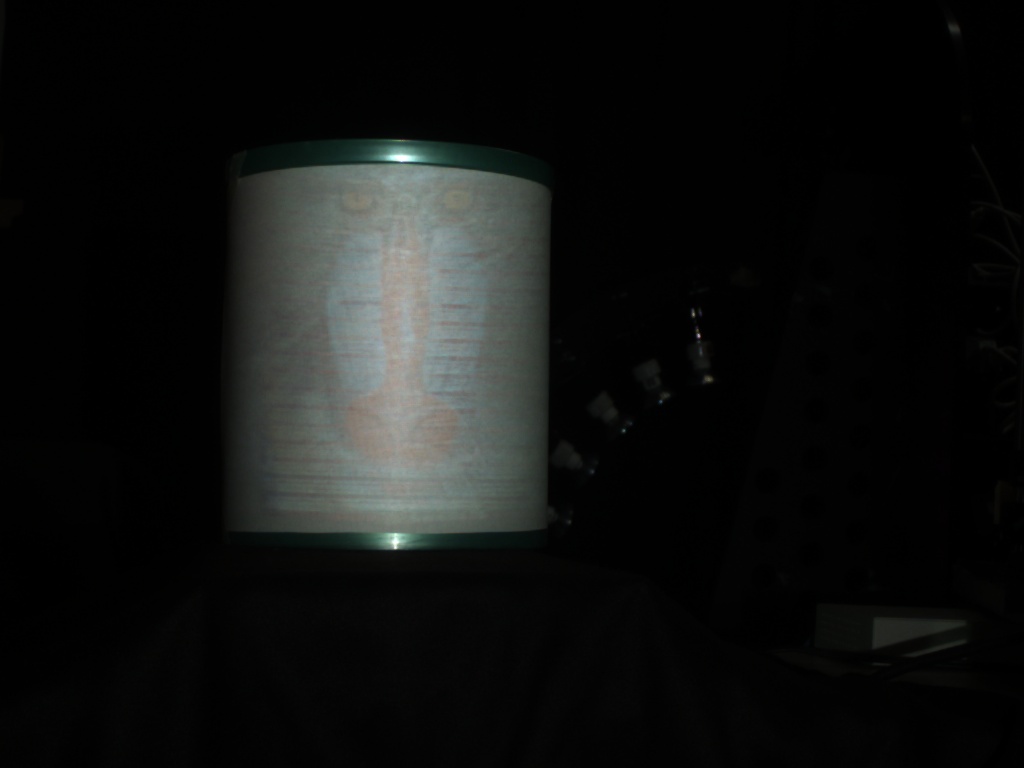}
%    \FramedBox{3.5cm}{1\textwidth}{Plane result with dense depth1.}
\caption{}
\end{subfigure}
\begin{subfigure}{0.24\textwidth}
    \includegraphics[trim={0cm 2.7cm 3cm 0cm},clip = true,width = 1\textwidth]{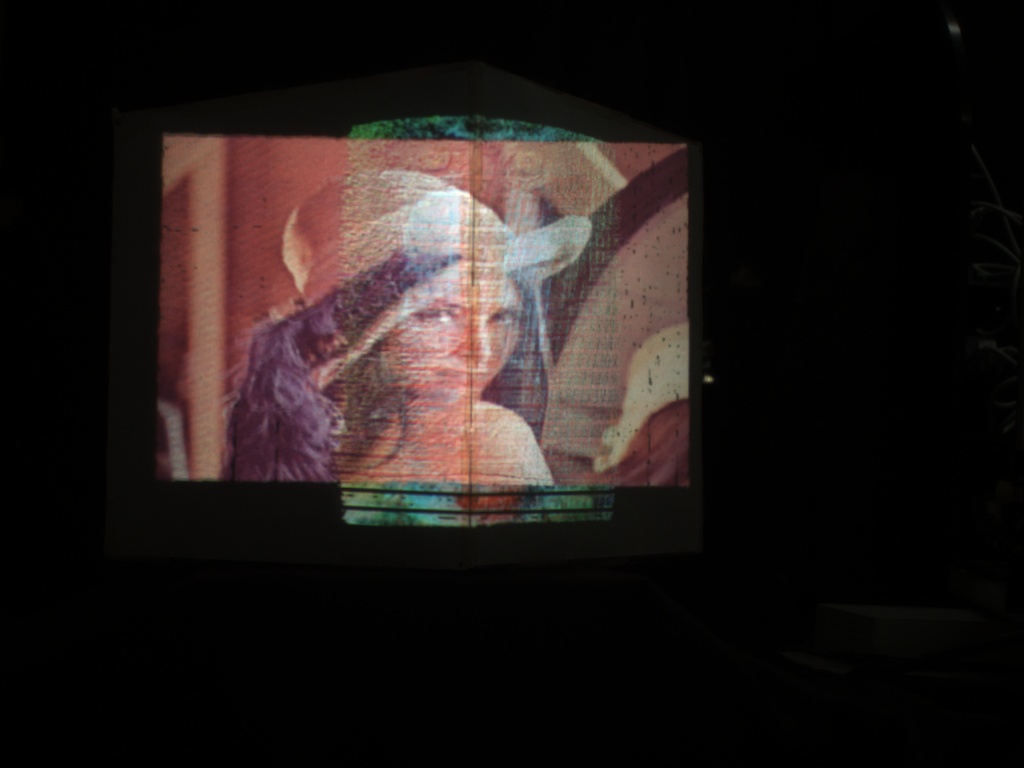}
%    \FramedBox{3.5cm}{1\textwidth}{Plane result with sparse depth2.}
\caption{}
\end{subfigure}
\begin{subfigure}{0.24\textwidth}
    \includegraphics[trim={4cm 4cm 6cm 4cm}, clip = true,width = 1\textwidth]{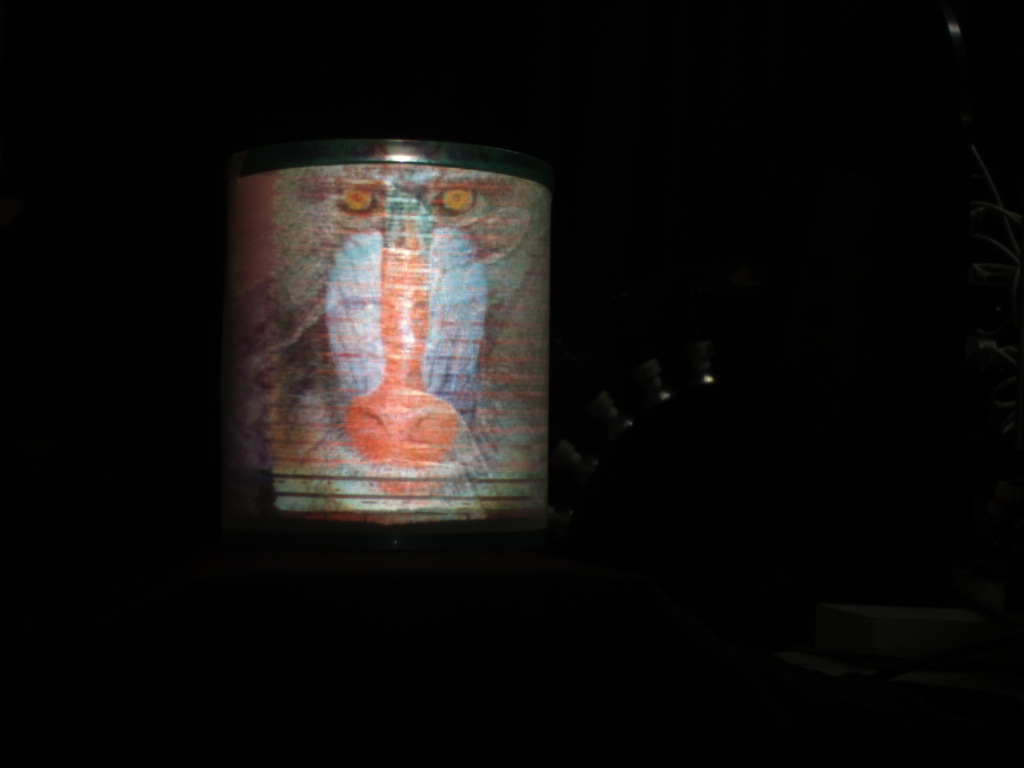}
 %   \FramedBox{3.5cm}{1\textwidth}{Plane result with dense depth2.}
\caption{}
\end{subfigure}
%\knote{[Hirukawa] input images for comparison of sparse and dense version.}
%
\begin{subfigure}{0.23\textwidth}
\includegraphics[trim={8cm 1cm 8cm 12cm},clip=true,width=1\textwidth]{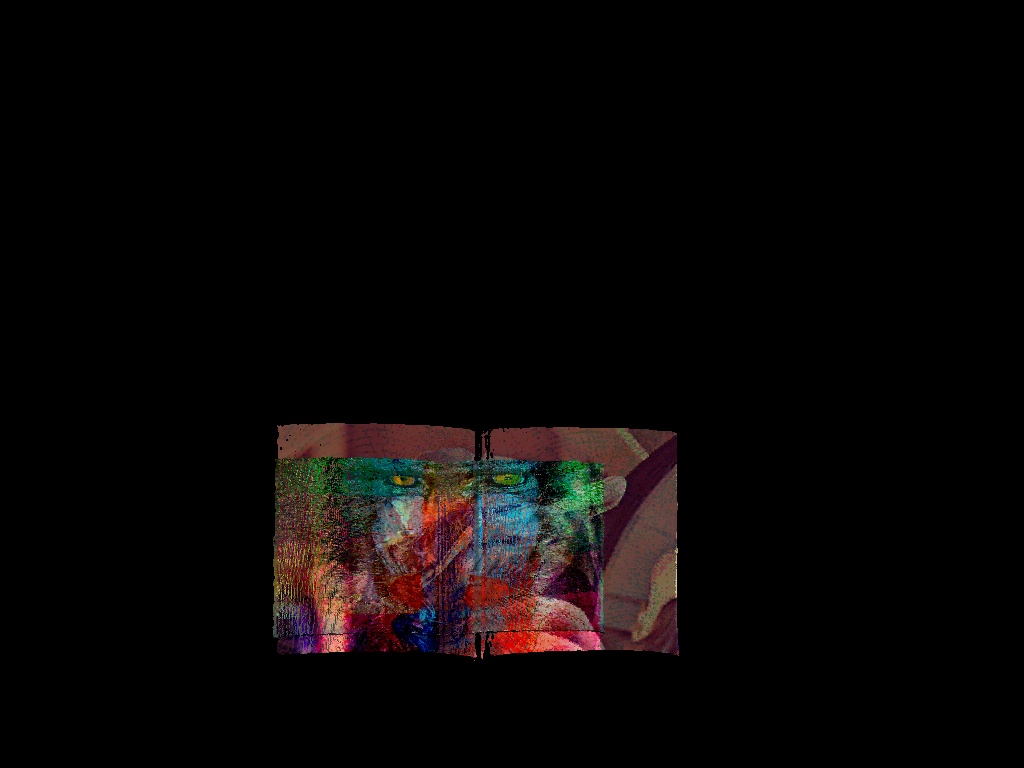}
\caption{}
\label{fig:patterne}
\end{subfigure}
\begin{subfigure}{0.23\textwidth}
\includegraphics[trim={8cm 7cm 8cm 6cm},clip=true,width=1\textwidth]{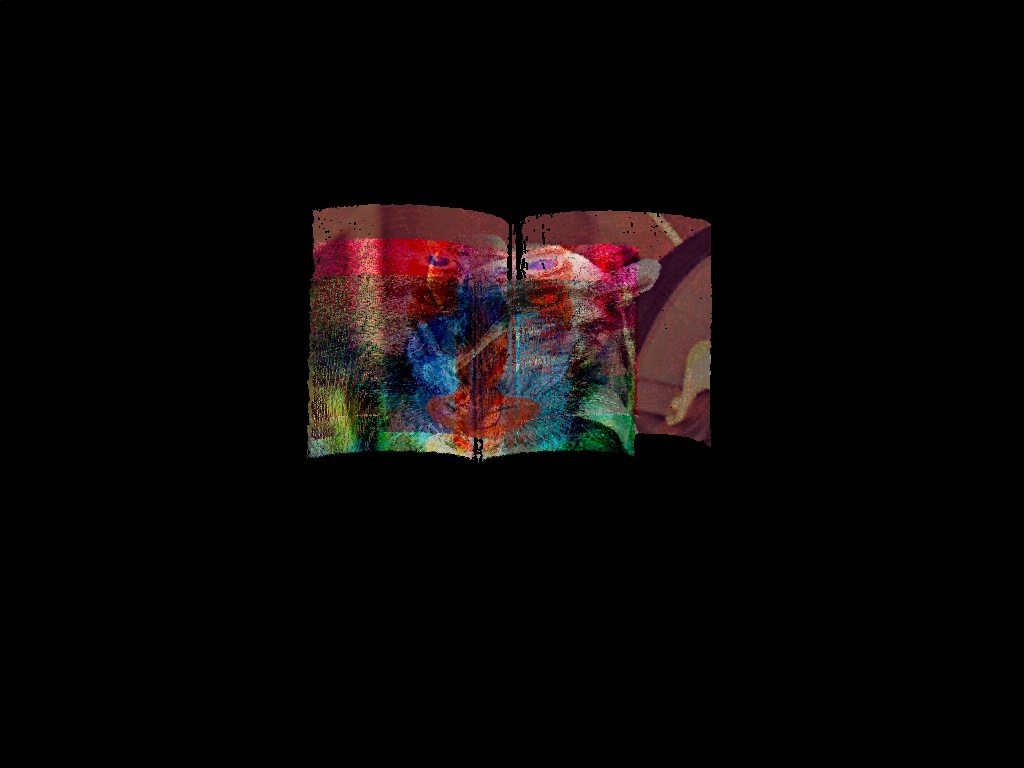}
\caption{}
\label{fig:patternf}
\end{subfigure}
\begin{subfigure}{0.23\textwidth}
\includegraphics[trim={5cm 0.5cm 5cm 9cm},clip=true,width=1\textwidth]{p_bilinear_pattern_1.jpg}
\caption{}
\label{fig:patterng}
\end{subfigure}
\begin{subfigure}{0.23\textwidth}
\includegraphics[trim={5cm 3.5cm 5cm 6cm},clip=true,width=1\textwidth]{p_bilinear_pattern_2.jpg}
\caption{}
\label{fig:patternh}
\end{subfigure}
\caption{\textit{Lena/Mandrill} pairs projected on non-planar scenes placed at different positions.
Projected patterns are generated with (a)(b)(e)(f) LF \cite{Scarzanella:psivt15} 
and (c)(d)(g)(h) the proposed $EO_0^{255}$ algorithm. (i)(j)(k)(l) are the 
    actual generated images for each projector in the two scenarios.} 
%\caption{\textit{Lena/Mandrill} pairs projected on non-planar objects with (a),(b) the LF \cite{Scarzanella:psivt15} and (c),(d) the proposed $EO_0^{255}$ algorithm.} 
\label{Fig_synthResults3D}
\end{figure}
\begin{figure}[t]
\centering
\begin{subfigure}{0.22\textwidth}
\includegraphics[trim={5cm 0cm 4cm 2cm},clip=true, height = 20mm]{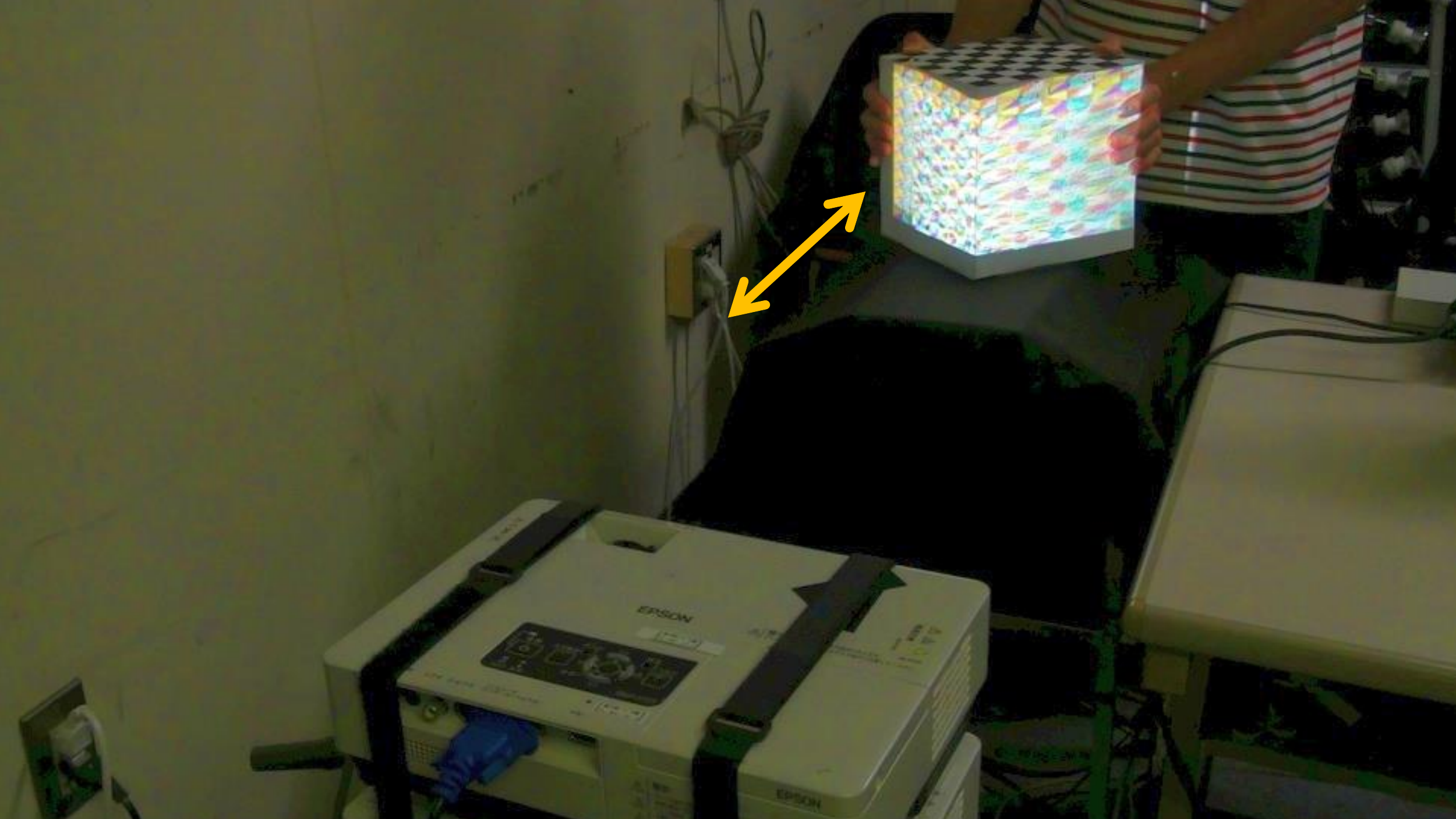}
\caption{}
\label{fig:positiona}
\end{subfigure}
\begin{subfigure}{0.22\textwidth}
\includegraphics[clip=true, height = 20mm]{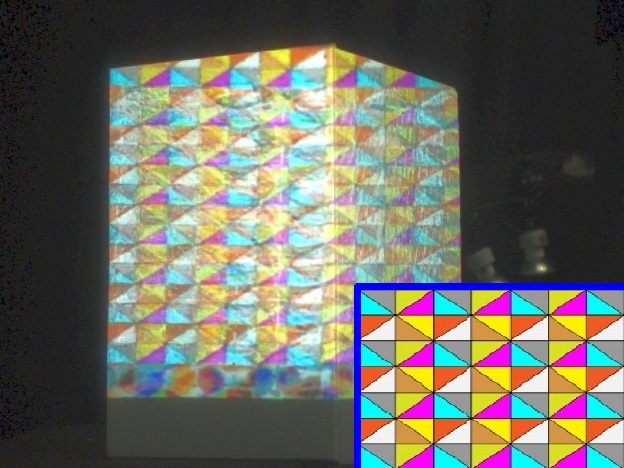}
\caption{}
\label{fig:positionb}
\end{subfigure}
\begin{subfigure}{0.22\textwidth}
\includegraphics[clip=true, height = 20mm]{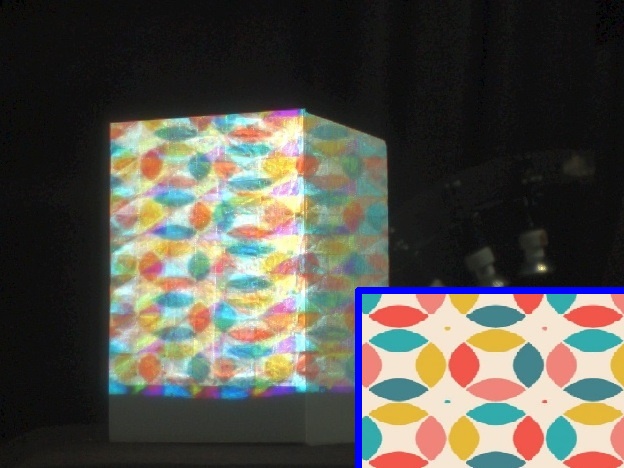}
\caption{}
\label{fig:positionc}
\end{subfigure}
\begin{subfigure}{0.24\textwidth}
\includegraphics [height = 20mm]{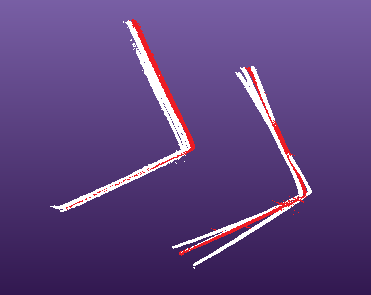}
\caption{}
\label{fig:positiond}
\end{subfigure}
%\knote{[Hirukawa] change for box.}
\vspace{-0.3cm}
\caption{(a) Test scene, projected patterns when the box is at the position (b) 
    nearer and (c) farther away from the projectors. In (b) and (c), the target 
    images for each object are shown inset at the bottom 
    right corner. (d) 3D-scanned boxes during the test (white) and ground 
    truth (red).}
\label{fig:positioning}
\vspace{-0.3cm}
\end{figure}

\section{Experiments}
\label{sec:experiment}

%\fnoteII{
%[Furukawa] Revise texts. Should consider what kind of experiments, should we do.
%Now we have 3D object version with sparse and dense calculation.
%It is better if we have some qualitative evaluation and/or comparison.
%Is there any meaning compared with PSIVT version?
%}

Our setup consists of two stacked EPSON LCD projectors ($1024 \times 768$ pixels) and a single CCD camera
(  $1600 \times 1200$ pixels)
as shown in Fig.~\ref{fig:config}. 
%The resolution of both the projectors is $1024 \times 
%768$ and camera is $1600 \times 1200$. 
%For calibration, we prepare a special 
%calibration toolbox where the Gray code pattern is synchronously projected by the projectors and captured by the CCD camera. 
We calibrate the system for both geometrically and photometrically, 
and 
%We 
test the proposed system under three tasks. 
%First, the image quality is numerically evaluated for various settings of our proposed algorithm against the Linear Factorization (LF) method in \cite{Scarzanella:psivt15}. 
First, the image quality of the proposed algorithm is numerically evaluated against the Linear Factorization (LF) method in \cite{Scarzanella:psivt15}. 
Second, the visual positioning system application scenario shown in Fig.~\ref{fig:app} is tested. 
Finally, the projection results on complex surfaces are shown 
%by testing the system 
in the scenario of virtual mask projection of Fig.~\ref{fig:basic}.

%We prepare several objects for target.

%For the image quality assessment, the patterns were projected on a matte plane placed on a motorised 
%stage for fine distance control at 80cm and 
%100cm, referred to as $D_1$ and $D_2$ respectively. The images used for our tests are the publicly available \textit{Lena}, \textit{Mandrill}, \textit{Peppers} and \textit{House}, stretched to the $1024 \times 768$ projector resolution.

\subsection{Image quality assessment}

We first assess the quality improvement of our dynamic range expansion 
technique using a planar screen and choosing the combinations 
%In order to objectively assess the quality of the generated images and compare 
%it with the state-of-the art, we considered the combinations 
\textit{Lena}/\textit{Mandrill}, \textit{Lena}/\textit{Peppers}, \textit{Peppers}/\textit{House} and 
\textit{Peppers}/\textit{Lena} for target images. The two screens were placed at approximately 80cm and 100cm from the projectors.
% in a $D_1/D_2$ configuration. 
For each combination, we 
projected the original image on the plane, and used it as a baseline for PSNR 
evaluation with our proposed method denoted as $EO^a_b$ where $(a,b)$ is the 
range of allowed intensity values, as well as the $LF$ 
method in \cite{Scarzanella:psivt15}. Sample results are shown in Fig. 
\ref{Fig_synthResults}, while exhaustive numerical results are given in Fig. 
\ref{Fig_realTableHist}. The results show 
that
the result images obtained by $EO_0^{255}$
have much wider dynamic range of colours 
than those of the results obtained by $LF$ and $EO_{-100}^{255}$.
However, in certain cases some artifacts 
are visible, even for results of $EO_0^{255}$.
%due to the impossibility of correctly generating patterns 
%recombining into images with highly contrasting colour characteristics. 
%that the increased 
%colour range required by the patterns computed with $LF$ and $EO_{-100}^{255}$ 
%generally results in greatly reduced contrast levels in the recombined images, 
%to the point that the images generated according to $LF$ are barely visible. On 
%the other hand, the images computed according to $EO_0^{255}$ exhibit contrast 
%levels close to the original images. However, in certain cases some artifacts 
%are visible.
%due to the impossibility of correctly generating patterns 
%recombining into images with highly contrasting colour characteristics. 
Interestingly, the nature of the artifacts is the same regardless of the method used, even though the artifacts might appear less pronounced due to the overall lower contrast of LF and $EO_{-100}^{255}$. 
This highlights the trade-off between artifacts, contrast levels and number of projectors in the scene. As part of our future work, we will investigate redundant systems with a higher number of projectors than targets to characterize better this trade-off. 
When comparing $EO_{0}^{255}$ with $EO_{-100}^{255}$, we can see that the latter suffers from drastically 
lower PSNR and SSIM levels due to the difference in image quality. The techniques were also qualitatively compared on non-planar objects as shown in Fig.~\ref{Fig_synthResults3D}, highlighting similar improvements in dynamic range when the proposed method is used.

\subsection{Visual positioning accuracy evaluation}

%\knote{Hirukawa-kun will do this experiment tomorrow. Here, several students 
%will set the object at the certain depth just using visual feedback. Then, we capture the Graycode and calculate RMSE for evaluation.}

As another application scenario, 
%we envisaged 
%is to 
we use the system to position objects at the right position and orientation based exclusively on visual feedback. Such system could be used both by human as well as robotic workers without extra sensors. For our tests, we asked several subjects to place a box in two predetermined positions using just the visual feedback from the projected pattern. The location of the box was then captured with a 3D scanner and compared with the ground truth position. 
The test scene, projected patterns and reconstructed 
shapes are shown in Fig~\ref{fig:positioning}.
The average RMSE was of 1.47\% and 1.26\% of the distance between 
the box and the projector for the positions closer and farther away from the projectors respectively. From the results, we can confirm that the proposed technique can be used 
for 3D positioning just using static passive pattern projectors.

%\knote{RMSE near 7.69mm、7.86mm、15.8mm -\>ave.10.45-\>1.47\%  (710), far 6.24, 
%11.8, 15.2 -\> ave. 11.08 -\> 1.26\% (880)}

\begin{figure}[t]
\centering
\begin{subfigure}{0.18\textwidth}
\includegraphics[width=.9\textwidth]{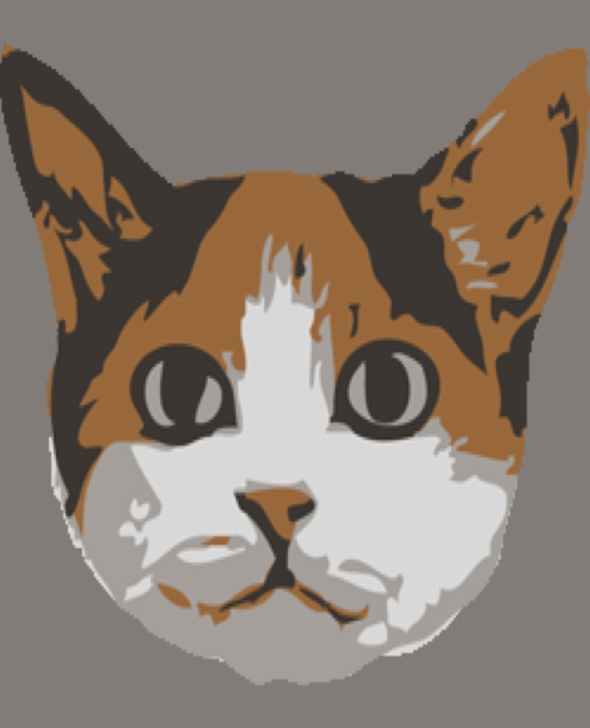}
%    \FramedBox{3.5cm}{1\textwidth}{Box result with dense depth1.}
\caption{}
\label{fig:origa}
\end{subfigure}
\begin{subfigure}{0.18\textwidth}
\includegraphics[width=.9\textwidth]{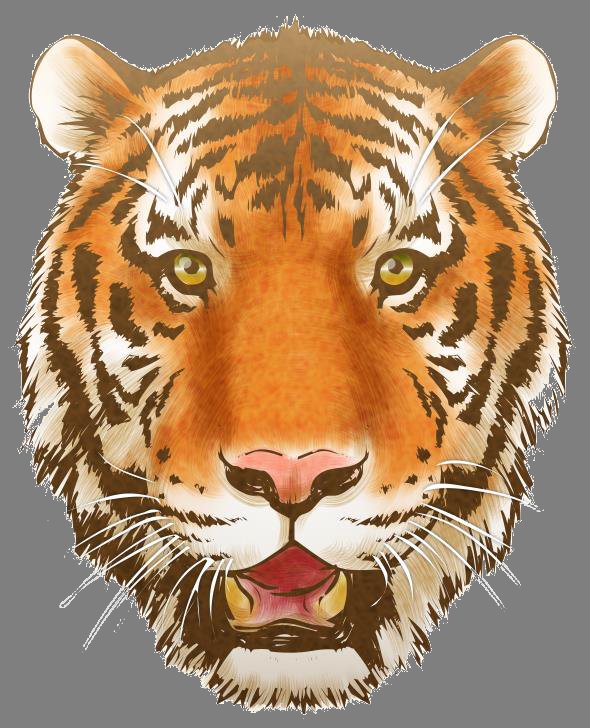}
%    \FramedBox{3.5cm}{1\textwidth}{Box result with dense depth1.}
\caption{}
\label{fig:origb}
\end{subfigure}
\begin{subfigure}{0.23\textwidth}
\includegraphics[trim={8cm 3.5cm 8cm 6cm},clip=true,width=1\textwidth]{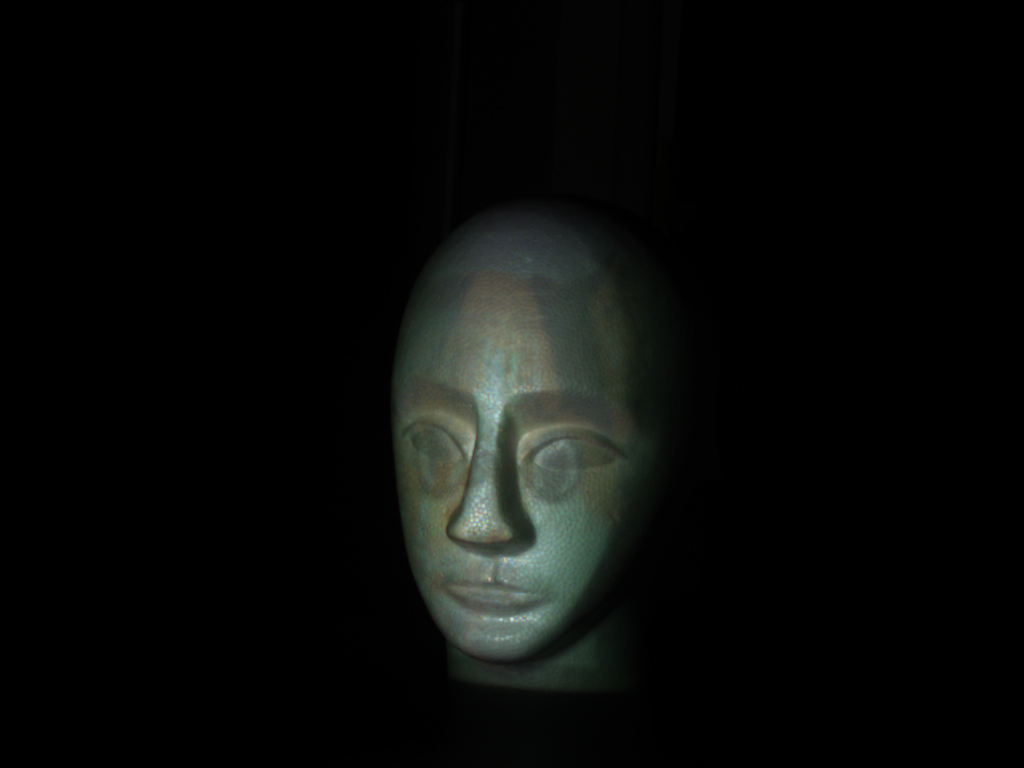}
%    \FramedBox{3.5cm}{1\textwidth}{Box result with dense depth2.}
\caption{}
\label{fig:sparseg}
\end{subfigure}
\begin{subfigure}{0.23\textwidth}
\includegraphics[trim={8cm 3.5cm 8cm 6cm},clip=true,width=1\textwidth]{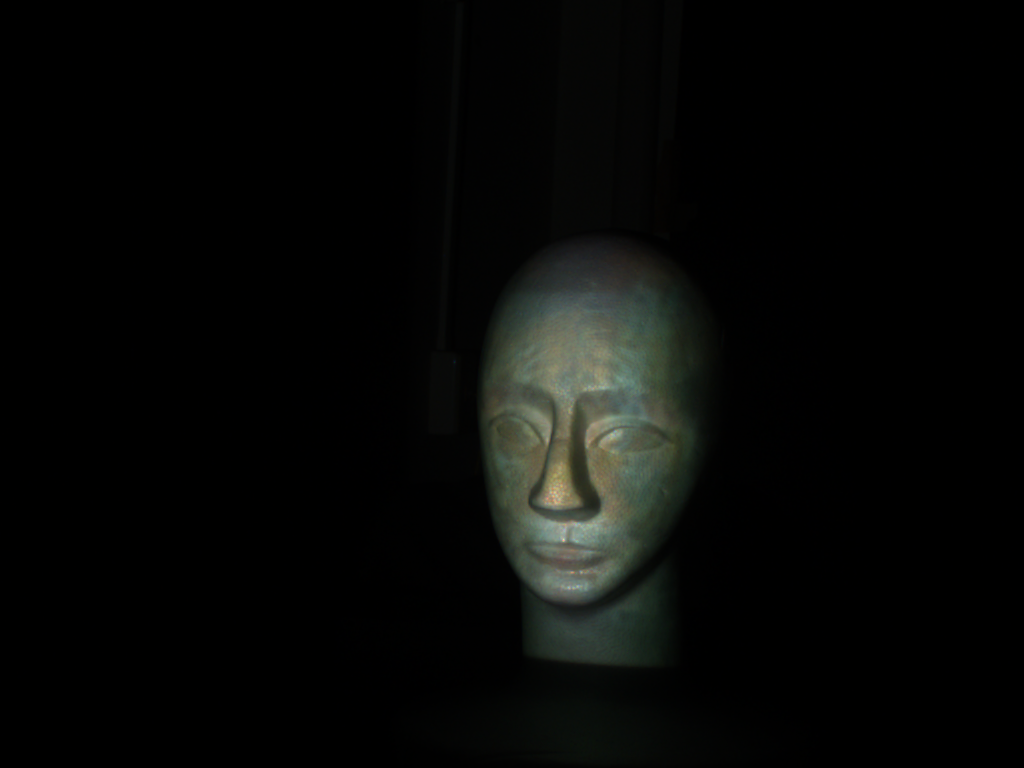}
%    \FramedBox{3.5cm}{1\textwidth}{Box result with sparse depth2.}
\caption{}
\label{fig:sparseh}
\end{subfigure}

%
%\begin{subfigure}{0.26\textwidth}
%\includegraphics[width=1\textwidth]{mannequin/gray_h.jpg}
%\caption{}
%\label{fig:grayc}
%\end{subfigure}
%
%\begin{subfigure}{0.26\textwidth}
%\includegraphics[width=1\textwidth]{mannequin/gray_v.jpg}
%\caption{}
%\label{fig:grayd}
%\end{subfigure}

%\knote{[Hirukawa] change for box.}
%\caption{Numerical evaluation of the proposed system. (a) Original 
%\textit{Peppers} image. (b) Recombined \textit{Peppers} image. (c) Recombined 
%pattern outside the predefined depths.}
%\vspace{-0.3cm}
%\caption{Multiple 3D objects projection results. (a) and (b) are results with our  method $EO_{0}^{255}$ at 
%    position 1 and  2, respectively and (c), (d) are results with $LF$ algorithm~\cite{Scarzanella:psivt15}. Camera settings are all the same.}
%\label{fig:comparison}

%\begin{subfigure}{0.23\textwidth}
%\includegraphics[trim={8cm 1.5cm 8cm 8cm},clip=true,width=1\textwidth]{mannequin/p_bilinear_pattern_1.jpg}
%\caption{}
%\label{fig:patterne}
%\end{subfigure}
%\begin{subfigure}{0.23\textwidth}
%\includegraphics[trim={8cm 3.5cm 8cm 6cm},clip=true,width=1\textwidth]{mannequin/p_bilinear_pattern_2.jpg}
%\caption{}
%\label{fig:patternf}
%\end{subfigure}
%
%
%\knote{[Hirukawa] change for box.}
%\vspace{-0.3cm}
%\caption{Projected patterns for (a) top and (b) bottom projectors with our 
%    method $EO_{0}^{255}$ and (c), (d) with $LF$ algorithm~\cite{Scarzanella:psivt15}.}
%\label{fig:pattern}
%\end{figure}
%
%\begin{figure}
\begin{subfigure}{0.23\textwidth}
\includegraphics[trim={8cm 3.5cm 8cm 6cm},clip=true,width=1\textwidth]{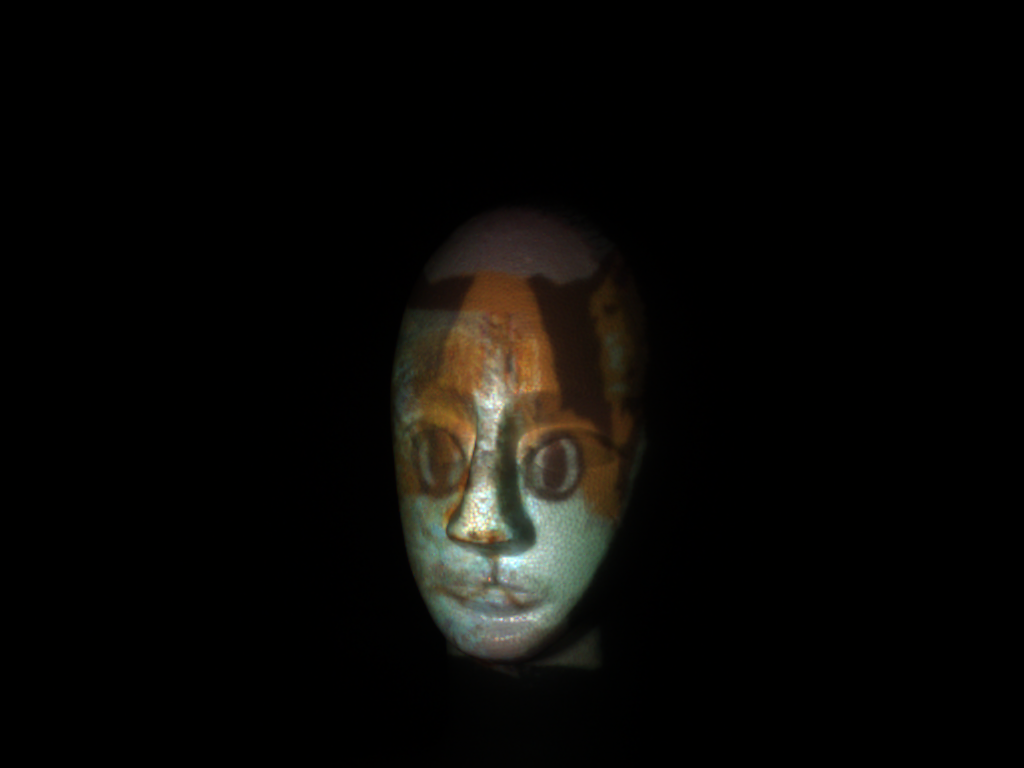}
%    \FramedBox{3.5cm}{1\textwidth}{Box result with dense depth1.}
\caption{}
\label{fig:densei}
\end{subfigure}
\begin{subfigure}{0.23\textwidth}
\includegraphics[trim={8cm 3.5cm 8cm 6cm},clip=true,width=1\textwidth]{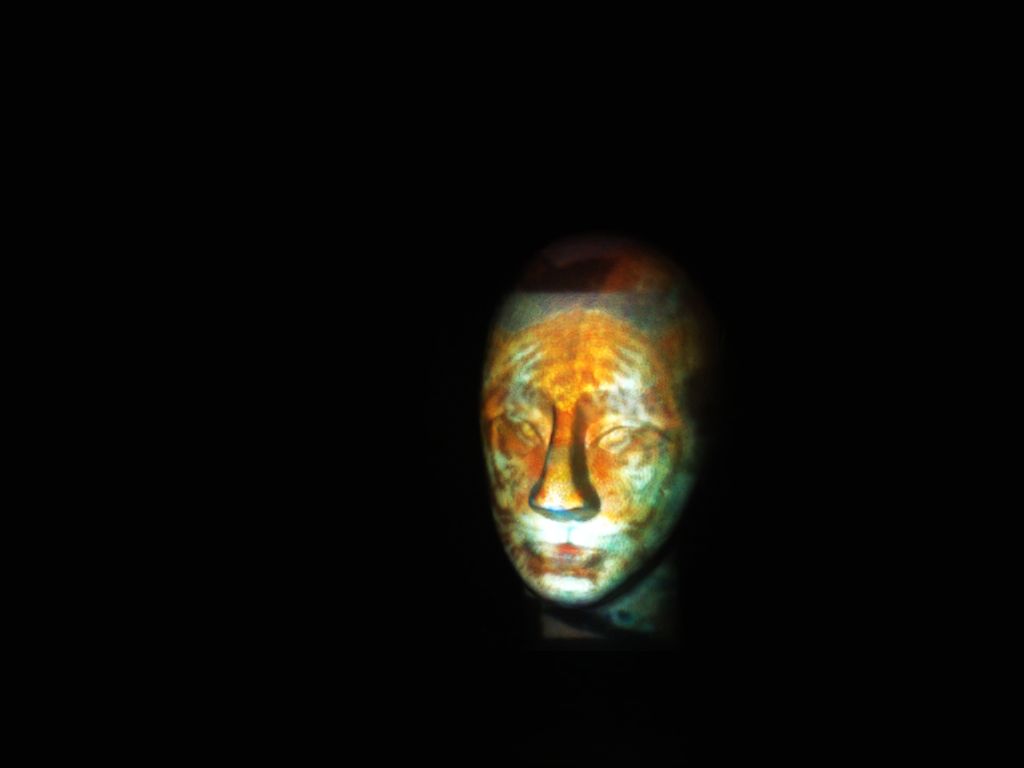}
%    \FramedBox{3.5cm}{1\textwidth}{Box result with dense depth2.}
\caption{}
\label{fig:densej}
\end{subfigure}
\begin{subfigure}{0.23\textwidth}
\includegraphics[trim={0cm 3cm 7.8cm 0cm},clip=true,width=1\textwidth]{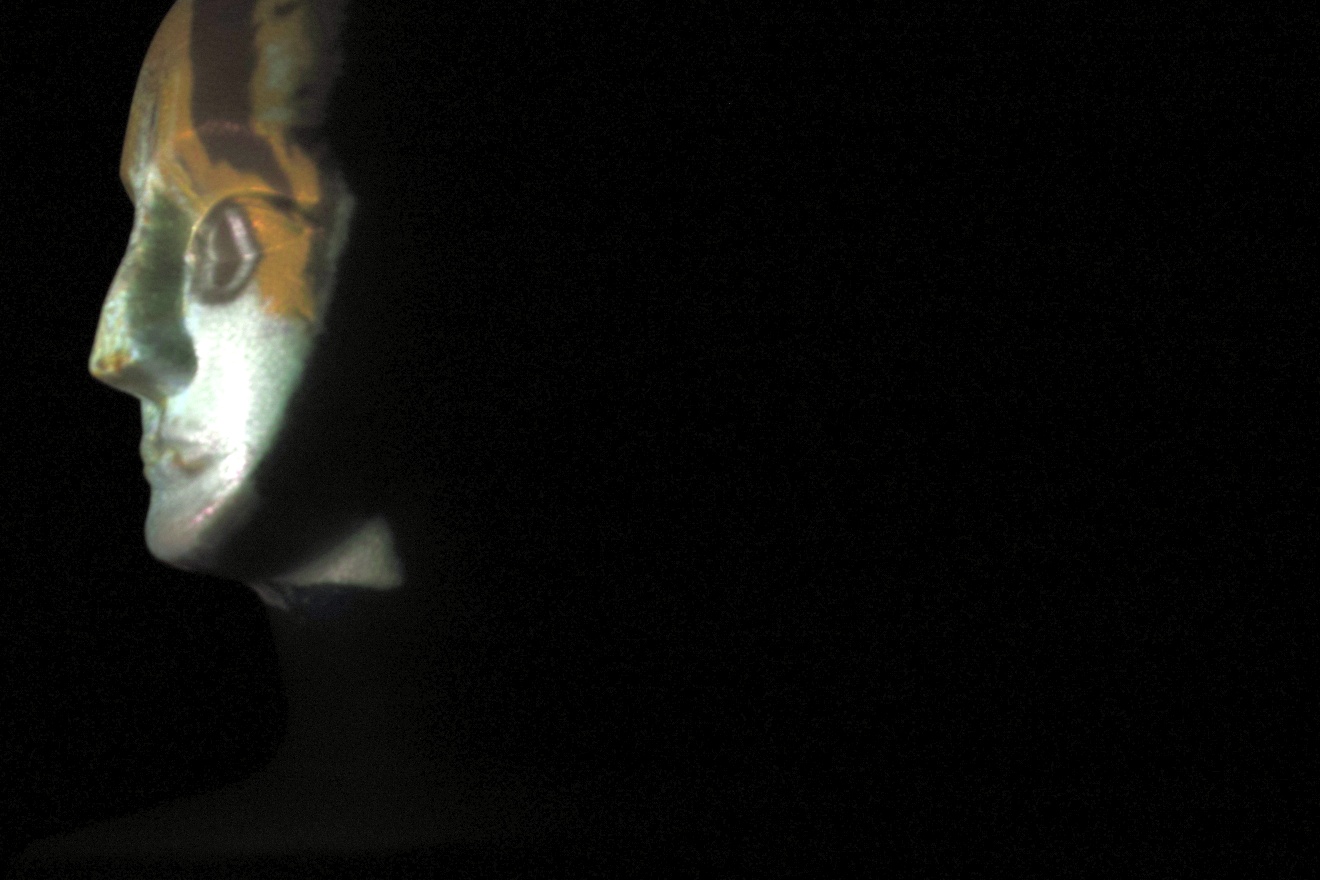}
%    \FramedBox{3.5cm}{1\textwidth}{Box result with sparse depth1.}
\caption{}
\label{fig:densek}
\end{subfigure}
\begin{subfigure}{0.23\textwidth}
\includegraphics[trim={3.9cm 3cm 3.9cm 0cm},clip=true,width=1\textwidth]{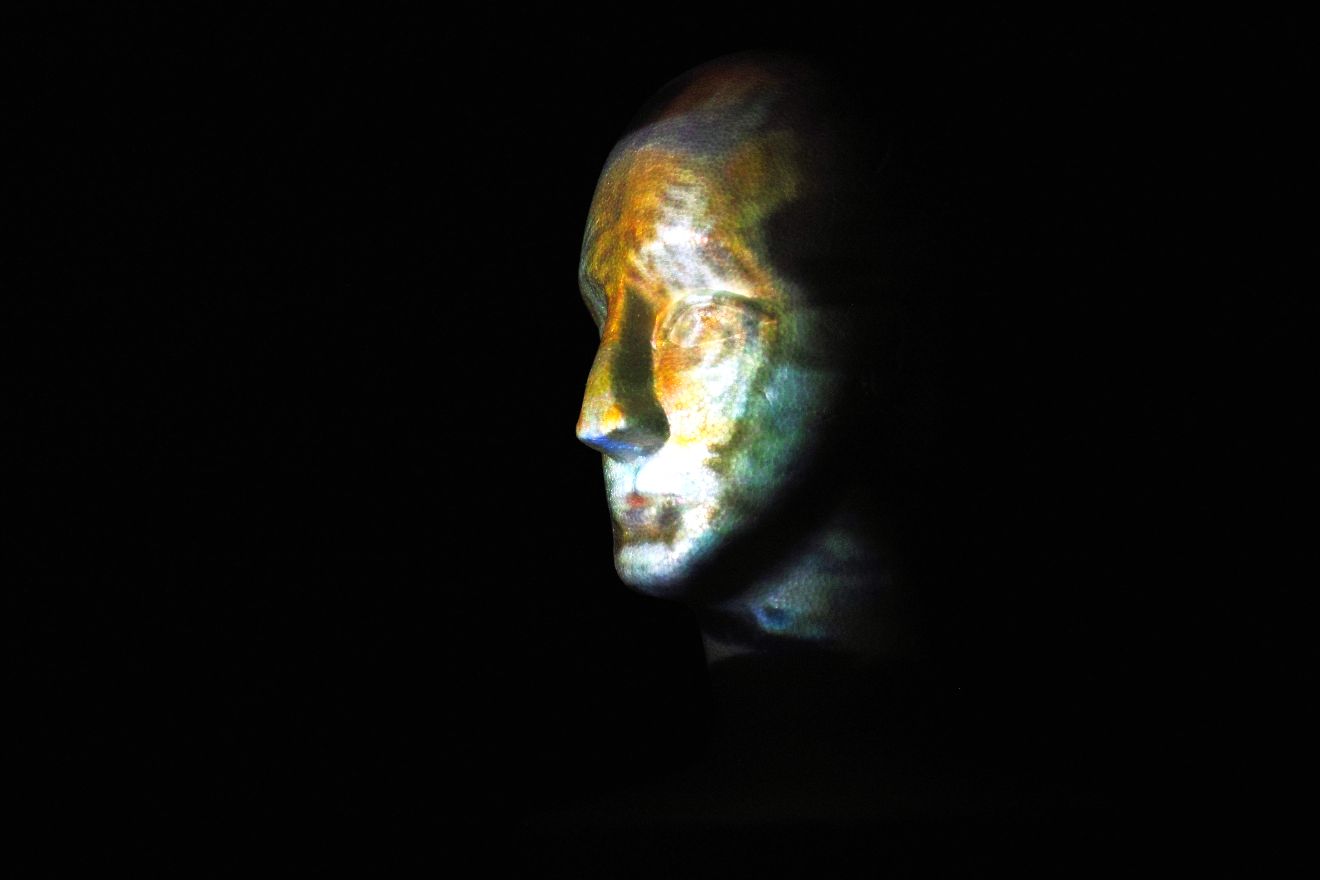}
%    \FramedBox{3.5cm}{1\textwidth}{Box result with sparse depth2.}
\caption{}
\label{fig:densel}
\end{subfigure}
%\knote{[Hirukawa] change for box.}
%\caption{Numerical evaluation of the proposed system. (a) Original 
%\textit{Peppers} image. (b) Recombined \textit{Peppers} image. (c) Recombined 
%pattern outside the predefined depths.}
\vspace{-0.3cm}
\caption{Projection results on objects with complex geometry. (a) and (b) are the target 
    images for two positions, (g), (f) are results with the $LF$ algorithm~\cite{Scarzanella:psivt15},
and
(i) to (l) are results with our  method $EO_{0}^{255}$ at 
    positions 1 and  2, respectively.}
\label{fig:comparison}

\end{figure}

\subsection{Independent image projection on multiple 3D objects}
%\knote{[Kawasaki] new experiment with Box.}
%\knote{We change the objects to 1.Mannequin and 2.Box+Cylinder case.}

For our third experiment, the system was tested on 3D objects with a more complex geometry, such as a mannequin head, as well as the combination of a square box 
and cylinders for the two scenarios mentioned in the introduction. 

In the first case, we projected the virtual masks in Fig.~\ref{fig:origa} and \ref{fig:origb} onto a mannequin placed at two 
different positions. Fig.~\ref{fig:sparseg} and \ref{fig:sparseh} show the results of $LF$ and 
Fig.~\ref{fig:densei} to Fig.~\ref{fig:densel} show the results of $EO_{0}^{255}$.
The figures show that the two images projected on the mannequin are clearly visible from all angles. Moreover, our proposed optimization significantly improves the result of $LF$. %Again, all the projected 
%patterns are static and just interference of them realizes such depth dependent projection.
%Further our dynamic range expansion is effectviely works for this case also.

\begin{figure}[t]
\centering
\begin{subfigure}{0.22\textwidth}
\includegraphics[trim={7cm 0cm 7cm 9cm}, clip = true, height = 20mm]{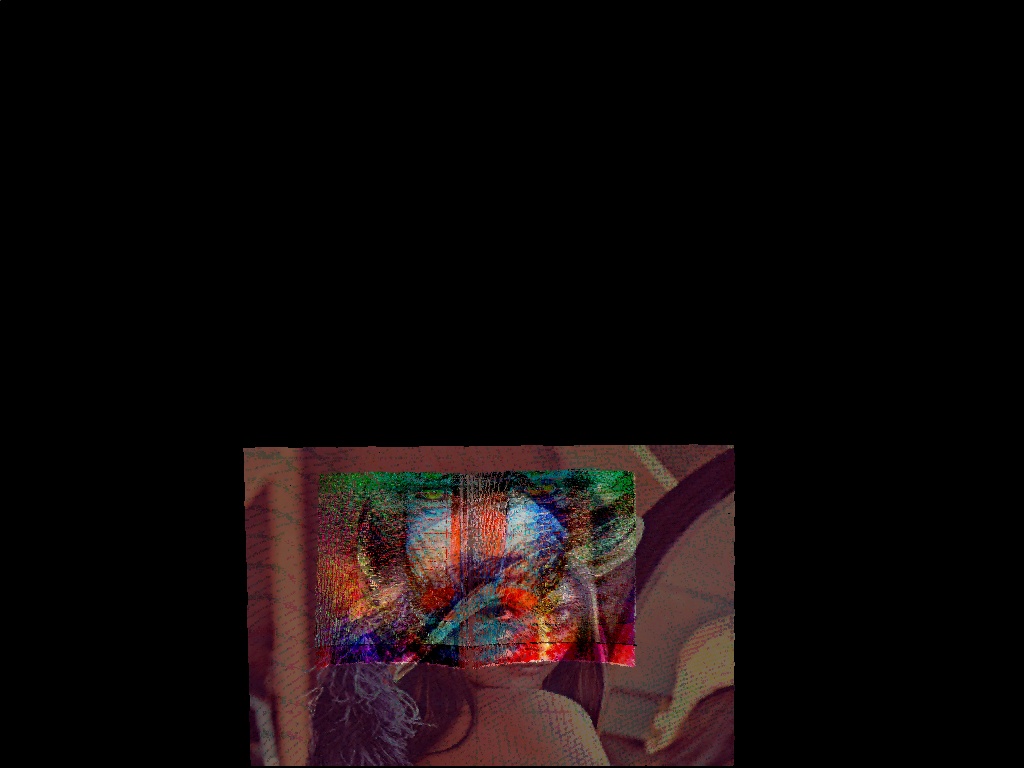}
%    \FramedBox{3.5cm}{1\textwidth}{Box result with dense depth1.}
\caption{}
\label{fig:boxplane_pa}
\end{subfigure}
\begin{subfigure}{0.22\textwidth}
\includegraphics[trim={7cm 2cm 7cm 7cm}, clip = true,height = 20mm]{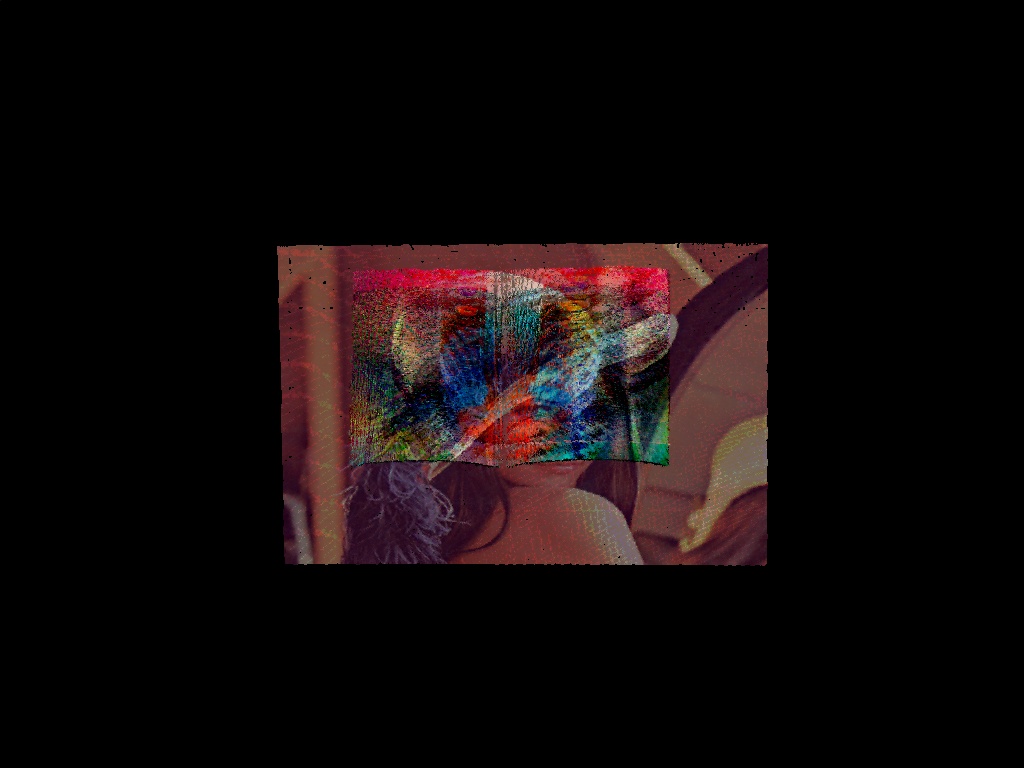}
%    \FramedBox{3.5cm}{1\textwidth}{Box result with dense depth2.}
\caption{}
\label{fig:boxplane_pb}
\end{subfigure}
\begin{subfigure}{0.22\textwidth}
\includegraphics[trim={5cm 0cm 3cm 0cm}, clip = true,height = 20mm]{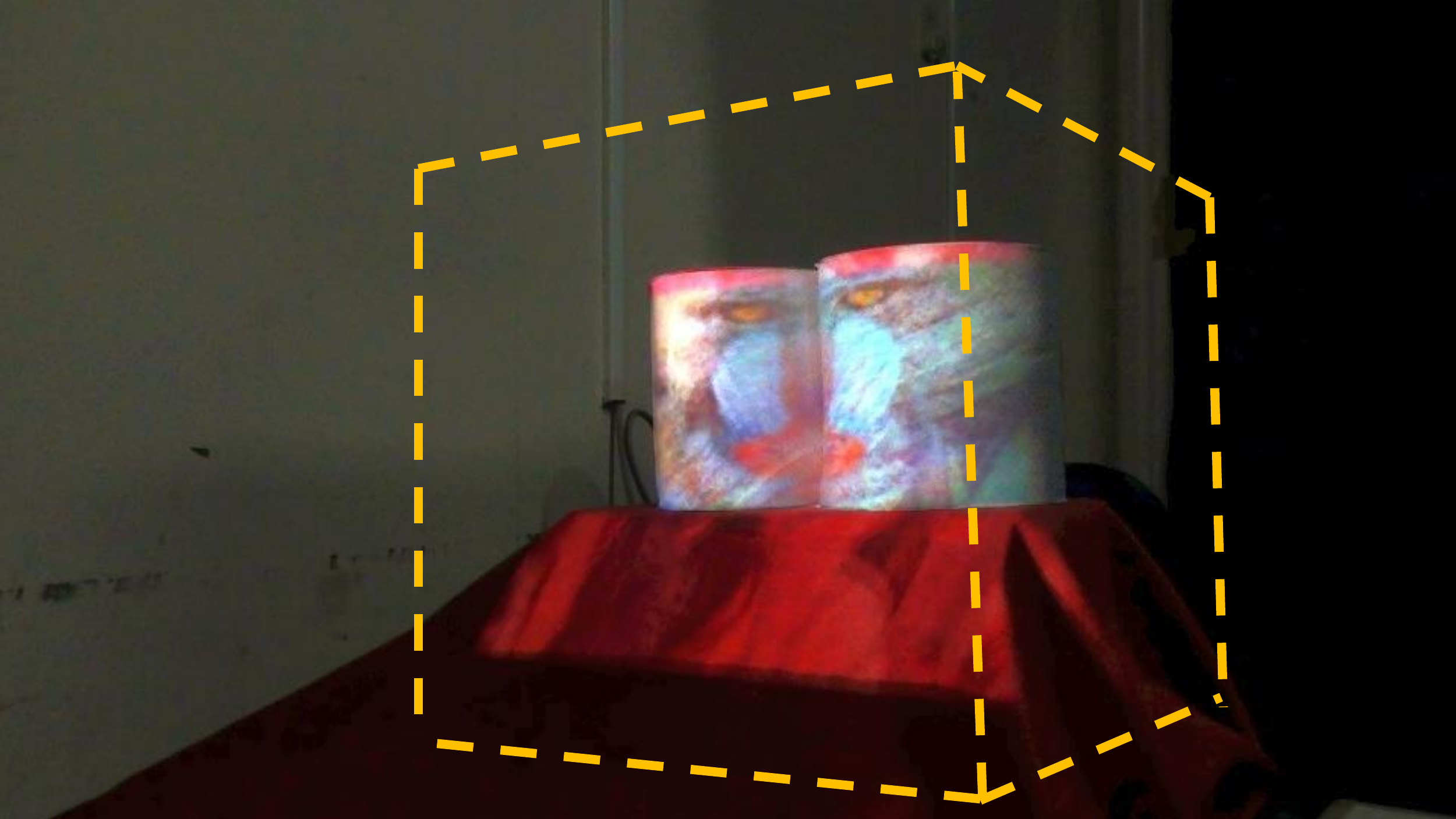}
%    \FramedBox{3.5cm}{1\textwidth}{Box result with sparse depth1.}
\caption{}
\label{fig:boxplane_pc}
\end{subfigure}
\begin{subfigure}{0.22\textwidth}
\includegraphics[trim={5cm 0cm 3cm 0cm}, clip = true,height = 20mm]{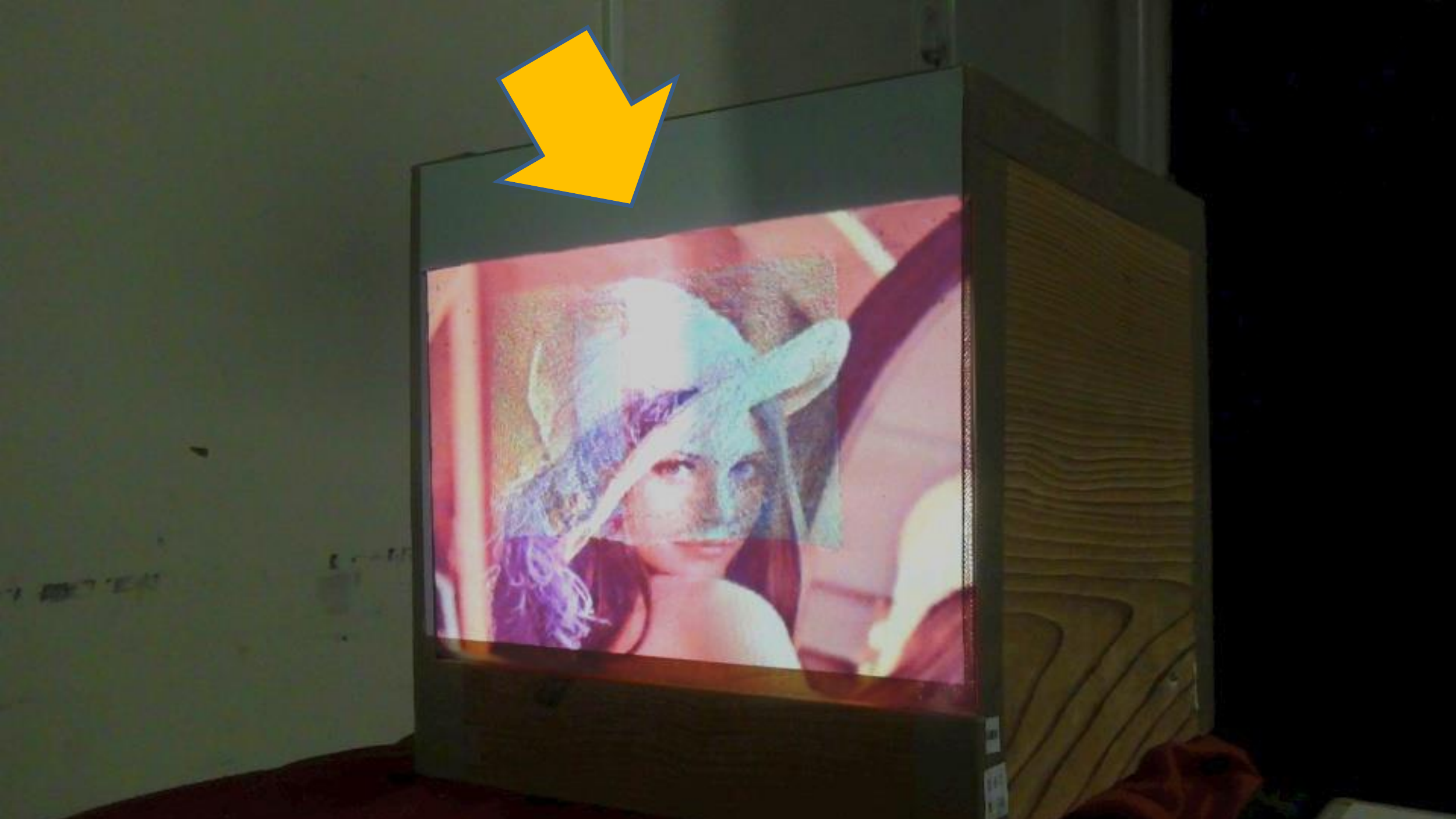}
%    \FramedBox{3.5cm}{1\textwidth}{Box result with sparse depth2.}
\caption{}
\label{fig:boxplane_pd}
\end{subfigure}
%\knote{[Hirukawa] change for box.}
%\caption{Numerical evaluation of the proposed system. (a) Original 
%\textit{Peppers} image. (b) Recombined \textit{Peppers} image. (c) Recombined 
%pattern outside the predefined depths.}
\vspace{-0.3cm}
\caption{Projected patterns for (a) top and (b) bottom projectors with our 
    method $EO_{0}^{255}$ and (c), (d) are the projected results.}
\label{fig:boxplane_pattern}

\end{figure}

Finally, we show how the system can be used for the object assembly workflow shown in Fig.~\ref{fig:app}.
Fig.~\ref{fig:boxplane_pa} and \ref{fig:boxplane_pb} are the calculated patterns 
with $EO_{0}^{255}$ for the two projectors, Fig.~\ref{fig:boxplane_pc} is the 
projected image on two cylinders and Fig.~\ref{fig:boxplane_pd} is the projected 
image on the large box placed outside the cylinders. We can confirm 
\textit{Lena/Mandrill} is clearly shown on each object, confirming that the technique has a potential to be used for correct positioning during object assembly.

%\begin{figure}
%\centering
%\begin{subfigure}{0.45\textwidth}
%\includegraphics[width=1\textwidth]{twoscreens1.jpg}
%\caption{}
%\end{subfigure}
%\begin{subfigure}{0.45\textwidth}
%\includegraphics[width=1\textwidth]{twoscreens2.jpg}
%\caption{}
%\end{subfigure}
%\caption{Prototype showing two images simultaneously projected on a matte and 
%semi-transparent screen for (a) \textit{Lena/Peppers} and (b) 
%\textit{Peppers/Lena}.}\label{fig:twoscreens}
%\end{figure}

\section{Conclusion} % \knote{Kawasaki\&Marco}}
\label{sec:conclusion}

%\knote{Should rewrite.}

In this paper, we propose a new pattern projection method which can simultaneously project independent images onto objects with complex 3D geometry at different positions. This novel system is 
realized by using multiple projectors
with geometrical calibration 
using a Gray code
%efficient corresponding points calculation method 
with a simple formulation to create 
suitable distributed interference patterns for each projector. In addition, an efficient 
calculation method including additional constraints of epipolar geometry to allow parallelisation and higher color dynamic 
range was proposed. Experiments 
showed the performance of a working prototype showing its improvement against the state of the art, and two application scenarios for object distance assessment and depth-dependent projection mapping. Our future work will concentrate on extending the system to more complicated scenes involving a higher number of projectors, and studying their optimality characteristics.
%Extensions will concentrate on increasing the dynamic range as well as 
%scaling the numbers of patterns and projectors in the prototype.
%calibration of multiple sets is planned.
%estimation of a consistent parameter with optimization via bundle adjustment.  
%With this method, it realize accurate calibration with short period of time  even in a scene 
%where it is difficult to perform a hard calibration so far.
%In the future 

%\section*{Acknowledgments}
%This work was supported by The Japanese Foundation for the Promotion of Science, 
%Grant-in-Aid for JSPS Fellows no.26.04041.

\clearpage

%------------------------------------------------------------------------- 
%\subsection{Citations}

\bibliographystyle{../bib/splncs}
%\bibliography{../bib/IEEEabrv,refs,../bib/h-kawa,../bib/JabRef}
\bibliography{refs,../bib/h-kawa,../bib/JabRef}

%The list of references is headed ``References'' and is not assigned a
%number in the decimal system of headings. The list should be set in small print and placed at the end of your contribution, in front of the appendix, if one exists.
%
%Do not insert a page break before the list of references if the page is not completely filled. Citations in the text are with square brackets and consecutive numbers, such as \cite{Alpher02}, or \cite{Alpher03,Herman04}.

%References are listed in alphabetic order by the surname of the first author, or the identifying word (e.g., in case of a website). Have
%all anonymized references at the beginning of the list.

%here would be your acknowledgement (if any) in the final accepted paper

%===========================================================

%this would normally be the end of your paper, but you may also have an appendix
%within the given limit of number of pages
\end{document}